\documentclass[10.5pt,twocolumn,twoside]{IEEEtran}

\usepackage{cite}
\usepackage{amsmath}
\usepackage{amssymb}
\usepackage{fontenc}
\usepackage{color}

\newtheorem{theorem}{Theorem}

\usepackage{algorithm}
\usepackage{algorithmic}
\usepackage{subfigure}
\usepackage{stfloats}
\usepackage{graphicx}
\usepackage[caption=false,font=footnotesize]{subfig}
\usepackage{url}
\usepackage{booktabs}
\usepackage{balance}
\usepackage{units}
\newcommand{\tabincell}[2]{\begin{tabular}{@{}#1@{}}#2\end{tabular}}
\usepackage{multirow}
\usepackage{threeparttable}
\usepackage{amsmath}
\allowdisplaybreaks[4]
\usepackage{xcolor}

\newenvironment{sequation}{\small\begin{equation}}{\end{equation}}

\begin{document}

\title{Learning for Cross-Layer Resource Allocation in MEC-Aided Cell-Free Networks}


\author{Chong~Zheng,~\IEEEmembership{Student~Member,~IEEE},
		Shiwen~He,~\IEEEmembership{Member,~IEEE}
        Yongming~Huang,~\IEEEmembership{Senior Member,~IEEE},
        and~Tony~Q.~S.~Quek,~\IEEEmembership{Fellow,~IEEE}
%

%
\thanks{C.~Zheng, and Y.~Huang are with the National Mobile Communications Research Laboratory, School of Information Science and Engineering, Southeast University, Nanjing 210096, China, and also with the Purple Mountain Laboratories, Nanjing 211111, China (e-mail: \{czheng; huangym\}@seu.edu.cn).}

\thanks{S. He is with the School of Computer Science and Engineering, Central South University, Changsha 410083, China. S. He is also with the Purple Mountain Laboratories, Nanjing 210096, China. (email: shiwen.he.hn@csu.edu.cn).}

\thanks{T.~Q.~S.~Quek is with the Information System Technology and Design Pillar, Singapore University of Technology and Design, Singapore 487372 (e-mail: tonyquek@sutd.edu.sg).}
}


\maketitle
\begin{abstract}
Cross-layer resource allocation over mobile edge computing (MEC)-aided cell-free networks can sufficiently exploit the transmitting and computing resources to promote the data rate. However, the technical bottlenecks of traditional methods pose significant challenges to cross-layer optimization. In this paper, joint subcarrier allocation and beamforming optimization are investigated for the MEC-aided cell-free network from the perspective of deep learning to maximize the weighted sum rate. Specifically, we convert the underlying problem into a joint multi-task optimization problem and then propose a centralized multi-task self-supervised learning algorithm to solve the problem so as to avoid costly manual labeling. Therein, two novel and general loss functions, i.e., negative fraction linear loss and exponential linear loss whose advantages in robustness and target domain have been proved and discussed, are designed to enable self-supervised learning. Moreover, we further design a MEC-enabled distributed multi-task self-supervised learning (DMTSSL) algorithm, with low complexity and high scalability to address the challenge of dimensional disaster. Finally, we develop the distance-aware transfer learning algorithm based on the DMTSSL algorithm to handle the dynamic scenario with negligible computation cost. Simulation results under $3$rd generation partnership project 38.901 urban-macrocell scenario demonstrate the superiority of the proposed algorithms over the baseline algorithms.
\end{abstract}


\bigskip

\section{Introduction}
\label{sec1}

With the rapid proliferation of the fifth generation (5G) communication technologies in recent years, people's life experience as well as industrial efficiency have achieved a further leap. However, to satisfy future demands for information and communications technology (ICT) in 2030, researchers in both academia and industry have shifted their attention to sixth generation (6G) communication systems \cite{Jiang21,HeS22arx}. Numerous promising wireless techniques, e.g., massive multiple-input multiple-output (MIMO) \cite{HeS21}, mobile edge computing (MEC) \cite{Zheng22} and cell-free  \cite{AmmarHA22}, are expected to tackle ongoing challenges, e.g., enormous wireless traffic, massive user access, and ultra-reliable low-latency communication. CF-MIMO systems have abilities to sufficiently exploit transmitting resources in massive user networks to promote the data transmitting rate through coordinated multi-point joint transmission.

Orthogonal frequency division multiple access (OFDMA) as an effective multiplexing scheme has been adopted in the $3$rd generation partnership project (3GPP) \cite{ZhangY18}. The transmission scheme design in the OFDMA CF-MIMO system contains two crucial issues, i.e., subcarrier allocation (SA) at the media access control layer \cite{SunY17} and beamforming (BF) design at the the physical layer \cite{Zhang09}. These two aspects are cross-layer and tightly intertwined in the downlink transmission and thus need to be jointly optimized for better-transmitting performance. However, most relevant works optimize the SA and BF separately without any consideration of their coupled relations. Specifically, many works have investigated the separate SA optimization problem from the perspective of the traditional optimization \cite{WenD21}, game theory \cite{Chen20}, and machine learning \cite{ZhouY20}. On the contrary, many other literatures \cite{HuaM21,BaoX19,SongH21} study the BF design from the same three perspectives while completely ignoring the influence of the SA design. 
Although separately considering the SA optimization or the BF design in these mentioned works both show the promise of improving system performance, the separate optimization will cause significant performance loss or even become impractical due to the tightly coupled relationship between the SA and the BF in the OFDMA CF-MIMO systems. In addition, the power allocation (PA) as an important part of the transmission design has been jointly considered with the SA optimization or the BF design in many related works, i.e., \cite{WenD21,ShenHL22,FuY20,XiuY21}. We would like to point out that the PA is not the focus of this paper and thus the reviews of the works related to the joint PA and SA/BF will not be detailed here. Nevertheless, it should be emphasized that the design of the PA is implicit in the BF design in this paper.

In fact, researchers have long realized the importance of the joint SA and BF optimization for improving the transmission performance of wireless systems, and have conducted continuous research on this joint optimization problem \cite{ChenG07,PanY04,VahidniaR13,LiB21,ZhouJ22,ZhaiB21,Vars22}. However, the optimization variables are coupled in high-dimensional space when considering the joint optimization of the SA and BF, which poses tough challenges, e.g., high complexity, NP-hard, etc., to the solution of this optimization problem. Especially when the CF-MIMO system is further involved, the solution of this joint optimization problem will be more challenging. On the one hand, to simplify the problem, exisited works, i.e. \cite{ChenG07,PanY04,VahidniaR13,LiB21,ZhouJ22,ZhaiB21,Vars22}, consider the joint optimization of the SA and BF in problem formulation but roughly decouple the joint optimization problem into several subproblems and alternately optimize the SA and BF according to these subproblems. 
It is not difficult to observe that these alternately optimization methods based on the problem decoupling can reduce the difficulty in solving the joint optimization problem. Nevertheless, these decoupled methods have actually broken the coupled relationship between the optimization variables and technically belong to the separate optimization of the SA and BF.

On the other hand, some efforts from the perspective of the traditional optimization have also been made in \cite{SunY18,Bayat20,Zakeri21} to design the SA and BF in a joint optimization manner without decoupling operations. Specifically, the authors in \cite{SunY18} study the SA and BF joint optimization for multiple-input single-output (MISO) multicarrier non-orthogonal multiple access (MC-NOMA) systems, where the joint design of the SA and BF is optimized for maximization of the weighted system throughput. Then, the monotonic optimization, which entails a high computational complexity, is applied to yield the optimal solution to the joint optimization problem. Bayat et al. \cite{Bayat20} focus on the MISO heterogeneous network and propose a joint SA and BF scheme based on the semidefinite relaxation and the convex programming to maximize the sum rate of the network, where the inter-layer interference, minimum data rate, minimum harvested energy and maximum transmitting power are considered in the constraints. In \cite{Zakeri21}, the authors jointly design the SA and BF in a multiple antennas NOMA network aiming to maximize the worst-case energy efficiency. To handle the difficulty, the authors in \cite{Zakeri21} adopt the Dinkelbach algorithm to convert the formulated problem into an equivalent rank-constrained semidefinite programming (SDP) problem and then solve it using sequential fractional programming. Consequently, the superiority brought by the joint optimization of the SA and BF has been preliminarily demonstrated in the foregoing works. However, the investigations in \cite{SunY18,Bayat20,Zakeri21} also reveal many intractable difficulties, i.e., the non-convexity, the complexity disaster, the NP-hard issue, etc., faced by the traditional optimization methods in solving the joint SA and BF optimization problem.

Deep learning (DL) is viewed as a key enabler to break the technical bottlenecks of traditional methods in solving cross-layer optimization problems. Particularly, some recent works related to the cross-layer optimization, i.e., the joint user scheduling and BF design \cite{HeS22twc,HuangR22,BandiA20,HeS23twc}, joint base station clustering and BF design \cite{HeY20}, have investigated the learning-based joint optimization and demonstrated the potential of the DL in handling the cross-layer optimization problems. Unfortunately, to our best knowledge, few works have explored cross-layer optimization with respect to the joint SA and BF design from the perspective of DL. In fact, the joint SA and BF design is indeed more challenging than the other joint optimizations such as the joint user scheduling and BF design due to the fact that the latter can be regarded as a special case of the former under the single carrier condition. Moreover, many open challenges in the field of DL, i.e., the costly manual labeling, the fixed dimensions of the input and output of neural networks (NN), and the dimension disaster for massive output, also pose huge obstacles to the learning-based joint SA and BF optimization. For example, the difficulties of traditional methods in obtaining the optimal solution of the joint SA and BF optimization pose significant challenges to the label acquisition for the supervised learning based solutions. In addition, when the scale of the communication network increases, the dimension of the joint optimization variables, i.e., the joint SA and BF variable, will increase exponentially, which can cause a dimensional disaster for the NNs. Furthermore, the immutability of normal NNs in the dimensions of input and output limit the scalability and transferability of conventional DL methods in handling dynamic wireless network scenarios.

As for the MEC technique, it can contribute powerful distributed computing capabilities to the learning-based optimization algorithms within the radio access network by equipping the edge nodes, i.e., WiFi access point or small base station (SBS), with the processing servers \cite{Faraci20,ZhengC21,CaoX23}. In addition, the advantages of the MEC framework in distributed computing architecture can provide supporting infrastructure for improving the scalability and transferability of learning-based wireless transmission schemes from the perspective of distributed learning. Thereby, the MEC framework can serve as a promising paradigm for future intelligent wireless communication networks \cite{YangB22,ZhengcG21,YangB21}. In this paper, the joint SA and BF optimization for the MEC-aided multi-user multi-subcarrier OFDMA CF-MIMO communication system is studied from the perspective of DL to maximize the weighted sum rate while subjecting to the constraints of the maximum transmitting power per SBS as well as the minimum data rate per user. Specifically, our contributions are summarized as follows:
\begin{itemize}
	\item We study the joint SA and BF optimization in the MEC-aided CF-MIMO system from the perspective of DL and convert the underlying joint optimization problem into a joint multi-task problem. Then we design two novel and general loss functions, i.e., negative fraction linear (NFL) loss and exponential linear (EL) loss whose advantages have been proved by theorems and simulations, for the preparation of solving the joint optimization problem in a self-supervised manner so as to address the challenge of label acquisitions in learning-based joint SA and BF optimization. 
	\item  We propose a centralized multi-task self-supervised learning (CMTSSL) algorithm based on the designed losses and multi-task self-supervised learning to finally solve the joint optimization problem. The newly designed losses as well as the proposed CMTSSL algorithm are also applicable to any other optimization problems with the same general form as the problem formulation in this paper.
	\item We further design a distributed multi-task self-supervised learning (DMTSSL) algorithm to reduce the computational complexity as well as improve the system performance, while addressing the dimensional disaster faced by the centralized learning when the communication network scale increases.
	\item We further develop a distance-aware transfer learning (DATL) algorithm for the dynamic scenario where a new SBS is suddenly added to the communication system, such as emergency communication scenarios, unmanned aerial vehicle-assisted communication scenarios and etc., by exploiting the scalability of the proposed DMTSSL algorithm. The designed DATL algorithm realizes the model transfer by sensing the distances between each SBS, which is sample but can effectively handle the dynamic scenario with negligible cost.
\end{itemize}

The remainder of this paper is organized as follows. The system model and problem formulation are presented in Section II. In Section III, the solution methodology is presented with details. Then, simulation results are discussed in Section IV. Finally, conclusions are drawn in Section V.

\begin{figure}[!t]
\centering
\includegraphics[width=0.45\textwidth]{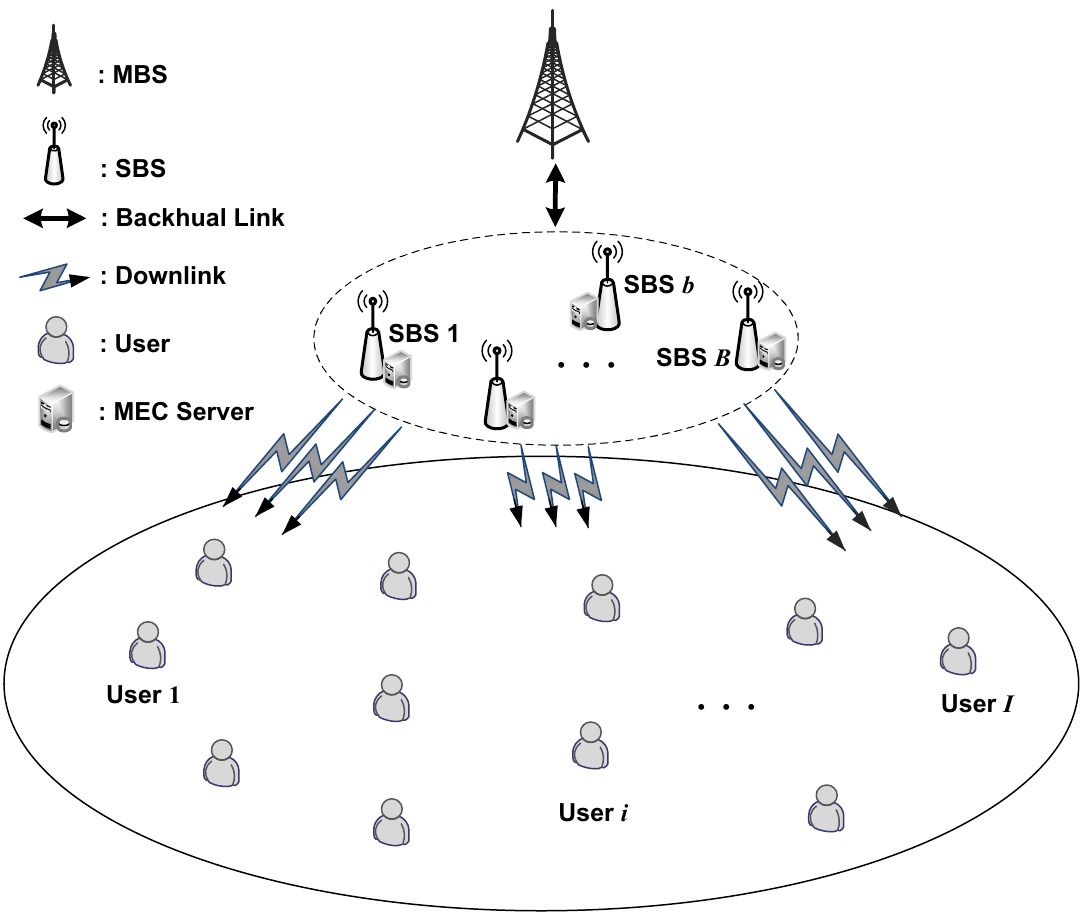}
\caption{Hierarchical architecture of the MEC-aided CF-MIMO system under investigation.}
\label{Sys}
\end{figure}

\section{SYSTEM MODEL AND PROBLEM FORMULATION}
\label{sec2}

\subsection{System Model}
We investigate the joint SA and BF for the coordinated multi-point joint downlink transmission of MEC-aided CF-MIMO communication system, including one macro base station (MBS), $B$ multi-antenna SBS and $I$ multi-antenna users, as illustrated in Fig. \ref{Sys}. Assume that each SBS is equipped with $M_{\rm t}$ transmitting antennas denoted as $\mathcal{M_{\rm t}}=\left\{ 1,2,\cdots,{M_{\rm t}}\right\}$ and each user is equipped with $M_{\rm r}$ receiving antennas denoted as $\mathcal{M_{\rm r}}=\left\{ 1,2,\cdots,{M_{\rm r}}\right\}$. Moreover, we also assumed that each SBS has $N$ subcarriers for the downlink transmission, denoted as $\mathcal{N}=\left\{ 1,2,\cdots,N\right\}$. The MEC server, which can be the general-purpose computer or server, is deployed on each SBS to provide certain computing and caching capabilities. For simplicity, let $\mathcal{B}=\left\{ 1,2,\cdots,B\right\}$ and $\mathcal{I}=\left\{ 1,2,\cdots,I\right\}$ be the set of SBSs and users, respectively. Let $\mathbf{v}^{b}=\left[\mathbf{v}_{1}^{b},\cdots,\mathbf{v}_{I}^{b}\right]\in\mathbb{C}^{N\times I}$ denote the SA indicator of the $b$-th SBS, where $\mathbf{v}_{i}^{b}=\left[v_{1,i}^{b},\cdots,v_{N,i}^{b}\right]^{\textrm{T}}\in\mathbb{C}^{N\times1}$ and $v_{n,i}^{b}$ is a binary variable that is $1$ if subcarrier $n$ of SBS $b$ is allocated to user $i$ and $0$ otherwise. The SA vector of all SBSs on subcarrier $n$ for user $i$ can be given by $\mathbf{v}_{n,i}=\left[v_{n,i}^{1},\cdots,v_{n,i}^{B}\right]^{\rm{T}}\in\mathbb{C}^{B\times1}$. Let $\mathbf{w}_{n,i}^{b}\in\mathbb{C}^{M_{\rm t}\times 1}$ represent the beamforming vector of SBS $b$ on subcarrier $n$ for user $i$ and $\mathbf{w}^{b}\in\mathbb{C}^{NM_{{\rm t}}\times I}$ denote the beamforming indicator of the $b$-th SBS. The beamforming vector of all SBSs on subcarrier $n$ for user $i$ can be denoted as $\mathbf{w}_{n,i}=\left[\left(\mathbf{w}_{n,i}^{1}\right)^{\textrm{T}},\cdots,\left(\mathbf{w}_{n,i}^{B}\right)^{\textrm{T}}\right]^{\textrm{T}}\in\mathbb{C}^{BM_{{\rm t}}\times1}$. Let $\mathbf{H}^{b}\in\mathbb{C}^{NM_{{\rm t}}\times IM_{{\rm r}}}$ denote the channel coefficient matrix between the $b$-th SBS and all users. $\mathbf{H}_{n,i}^{b}\in\mathbb{C}^{M_{\rm t}\times M_{\rm r}}$ represents the channel coefficient between user $i$ and SBS $b$ on subcarrier $n$. Then, the channel coefficient matrix between user $i$ and all SBSs on subcarrier $n$ can be denoted as $\mathbf{H}_{n,i}=\left[(\mathbf{H}_{n,i}^{1})^{\rm{T}},\cdots,(\mathbf{H}_{n,i}^{B})^{\rm{T}}\right]^{\rm{T}}\in\mathbb{C}^{BM_{{\rm t}}\times M_{{\rm r}}}$. For the convenience of implementation, we concatenate the channel coefficients of the system into a matrix $\mathbf{H}=\left[\left(\mathbf{H}^{1}\right)^{\rm{T}},\cdots,\left(\mathbf{H}^{B}\right)^{\rm{T}}\right]^{\rm{T}}\in\mathbb{C}^{BNM_{{\rm t}}\times IM_{{\rm r}}}$.

The baseband signal received by user $i$ on subcarrier $n$ can be defined by
\begin{sequation}
\begin{aligned}
\mathbf{y}_{n,i}=&\left(\mathbf{H}_{n,i}\right)^{\rm{H}}\left(\mathbf{\widetilde{v}}_{n,i}\mathbf{w}_{n,i}\right)s_{n,i}\\+&\sum_{j\neq i}\left(\mathbf{H}_{n,i}\right)^{\rm{H}}\left(\mathbf{\widetilde{v}}_{n,j}\mathbf{w}_{n,j}\right)s_{n,j}+\mathbf{z}_{n,i},
\end{aligned}	
\end{sequation}
\hspace{-5pt}where $s_{n,i}$ and $\mathbf{z}_{n,i}\in\mathbb{C}^{M_{{\rm r}}\times1}$ denote the baseband signal for user $i$ on subcarrier $n$ and the additive white Gaussian noise with $\mathcal{CN}\left(0,\sigma_{n,i}^{2}\right)$ at user $i$ on subcarrier $n$, respectively. $\mathbf{\widetilde{v}}_{n,i}\in\mathbb{C}^{BM_{{\rm t}}\times BM_{{\rm t}}}$ is the block diagonal matrix extended form $\mathbf{v}_{n,i}$.
\begin{sequation}
\mathbf{\widetilde{v}}_{n,i}=\textrm{diag}\left(v_{n,i}^{1}\text{,}\cdots,v_{n,i}^{B}\right)\otimes \mathbf{I}_{M_{\textrm{t}}},
\end{sequation}
\hspace{-4.5pt}where $\otimes$ represents the Kronecker Product. Then, the achievable rate of user $i$ is expressed as
\begin{sequation}
\begin{aligned}
r_{i}=\sum_{n=1}^{N}\textrm{log}_{2}\left|\mathbf{I}_{M_{\textrm{r}}}+\mathbf{S}_{n,i}\times\left(\mathbf{A}_{n,i}\right)^{-1}\right|,
\end{aligned}
\end{sequation}
\hspace{-5pt}where $\left|\cdot\right|$ represents the determinant of the matrix. $\mathbf{S}_{n,i}$ and $\mathbf{A}_{n,i}$ are the covariance matrix of the baseband signal and interference signal at user $i$ on subcarrier $n$, respectively, denoted as
\begin{sequation}
\begin{aligned}
\mathbf{S}_{n,i}=\left(\mathbf{H}_{n,i}\right)^{\textrm{H}}\left(\mathbf{\widetilde{v}}_{n,i}\mathbf{w}_{n,i}\right)\left(\mathbf{\widetilde{v}}_{n,i}\mathbf{w}_{n,i}\right)^{\textrm{H}}\mathbf{H}_{n,i},
\end{aligned}
\end{sequation}
\begin{sequation} \label{e1}
\begin{aligned}
\mathbf{A}_{n,i}=&\sum_{j\neq i}\left(\mathbf{H}_{n,i}\right)^{\textrm{H}}\left(\mathbf{\widetilde{v}}_{n,j}\mathbf{w}_{n,j}\right)\left(\mathbf{\widetilde{v}}_{n,j}\mathbf{w}_{n,j}\right)^{\textrm{H}}\mathbf{H}_{n,i}\\+&\sigma_{n,i}^{2}\mathbf{I}_{M_{\textrm{r}}}.
\end{aligned}
\end{sequation}

\subsection{Problem Formulation}
Under the investigated CF-MIMO communication scenario, the objective is to maximize the weighted sum rate of the system via joint SA and BF design. Accordingly, the optimization problem is formulated as
{\small \begin{subequations}\label{P1}
    \begin{align}
    P_1:\quad\quad
&\underset{\left\{\mathbf{w},\mathbf{\widetilde{v}}\right\}}{\max}\quad\sum_{i=1}^{I}\alpha_{i}r_{i},\label{P1a}\\
    {\rm{s.t.}}\quad \quad
    &r_{i}\geq r_{i}^{\textrm{min}},\forall i\in\mathcal{I},\label{P1b}\\
    &\sum_{n=1}^{N}\sum_{i=1}^{I}\left\Vert v_{n,i}^{b}\mathbf{w}_{n,i}^{b}\right\Vert _{2}^{2}\leq P_{\textrm{max}}^{b},\forall b\in\mathcal{B}, \label{P1c}
    \end{align}
\end{subequations}}
\hspace{-5pt}where $\mathbf{w}=\left[\mathbf{w}_{n,i}\right]_{i=1,n=1}^{I,N}$ and $\mathbf{\widetilde{v}}=\left[\mathbf{\widetilde{v}}_{n,i}\right]_{i=1,n=1}^{I,N}$ are the stack of the $\mathbf{w}_{n,i}$ and $\mathbf{\widetilde{v}}_{n,i}$, respectively. $\alpha_{i}$ denotes the prior of user $i$ and $r_{i}^{\textrm{min}}$ is the minimum data rate required by user $i$. $P_{\textrm{max}}^{b}$ represents the maximum allowable transmiting power of SBS $b$. Constraint \eqref{P1b} ensures that the minimum user rate demand of each user is to be satisfied. Constraint \eqref{P1c} prevents the transmitting power of each BS from the maximum allowable transmitting power. In general, the transmission scheme is designed with a separate consideration of SA and BF. Differently, in this paper, the designs of SA and BF are jointly optimized in the CF-MIMO communication system, which implies that the transmission scheme is jointly optimized across the physical layer and media access control layer for wireless communication systems.

It is not difficult to observe that problem \eqref{P1} is a mixed integer programming problem. To solve problem \eqref{P1}, the joint optimization of the SA and BF needs to be addressed, which is NP-hard \cite{Papad82}. Moreover, the achievable user rate $r_i$, constraint \eqref{P1b}, and constraint \eqref{P1c} are non-convex with respect to $\left\{\mathbf{w},\mathbf{\widetilde{v}}\right\}$ due to the interference channels in joint transmission systems. Accordingly, for problem \eqref{P1}, traditional optimization methods make it difficult to obtain the optimal solution, even the local optimal solution. Additionally, the relatively high computational complexity is another tough challenge for using traditional optimization methods to solve problem \eqref{P1} due to the fact that $\mathbf{w}$ and $\mathbf{\widetilde{v}}$ are in high dimensional and coupled spaces under the investigated CF-MIMO system. In the sequel, we focus on developing learning methods to solve problem \eqref{P1} while addressing the above technical challenges.

\subsection{Problem Recast}
To address the aforementioned technical challenges by developing learning methods, we convert problem \eqref{P1} into a multi-task learning problem whose learning tasks consist of the optimization objective and the constraints. Firstly, for simplicity and generalization, we herein introduce some auxiliary functions, i.e., $\mathbf{y}=\left\{ \mathbf{w},\mathbf{\widetilde{v}}\right\}$, $f\left(\mathbf{y},\mathbf{H}\right)=-\sum_{i=1}^{I}\alpha_{i}r_{i}$, $g_{i}\left(\mathbf{y},\mathbf{H}\right)=r_{i}^{\textrm{min}}-r_{i}$, and $l_{b}\left(\mathbf{y}\right)=\sum_{n=1}^{N}\sum_{i=1}^{I}\left\Vert v_{n,i}^{b}\mathbf{w}_{n,i}^{b}\right\Vert _{2}^{2}-P_{\textrm{max}}^{b}$, to equivalently rewrite problem \eqref{P1} as
{\small \begin{subequations}\label{P2}
    \begin{align}
    P_2:\quad\quad
&\underset{\mathbf{y}}{\min}\quad f\left(\mathbf{y},\mathbf{H}\right),\label{P2a}\\
    {\rm{s.t.}}\quad \quad
    &g_{i}\left(\mathbf{y},\mathbf{H}\right)\leq0,\forall i\in\mathcal{I},\label{P2b}\\
    &l_{b}\left(\mathbf{y}\right)\leq0,\forall b\in\mathcal{B}, \label{P2c}
    \end{align}
\end{subequations}}\vspace{-10pt}

Then, we view the optimization object as well as each constraint in problem \eqref{P2} as learning tasks, respectively. With respect to the optimization objective in problem \eqref{P2}, we can find from $f\left(\mathbf{y}\right)=-\sum_{i=1}^{I}\alpha_{i}r_{i}$ that this objective is the negative weighted sum rate which is expected to minimize to negative infinity. On this basis, the optimization objective \eqref{P2a} can be converted to a minimization learning task, formulated as 
{\small \begin{subequations}\label{task1}
\begin{align}
\underset{\mathbf{y}}{\min} \quad\quad
& f\left(\mathbf{y},\mathbf{H}\right),\\
    {\rm{s.t.}}\quad \quad
    &f\left(\mathbf{y},\mathbf{H}\right)\leq0.
\end{align}
\end{subequations}}
\hspace{-4pt}Similarly, in constraint (6b), $g_{i}\left(\mathbf{y},\mathbf{H}\right)$ is the residual between the tolerant user rate and the actual user rate, which is also expected to approach negative infinity. As such, for $\forall i\in\mathcal{I}$, constraint \eqref{P2b} can also be converted to a minimization learning task, expressed as
{\small \begin{subequations}\label{task2}
\begin{align}
\underset{\mathbf{y}}{\min} \quad\quad
& g_{i}\left(\mathbf{y},\mathbf{H}\right),\\
    {\rm{s.t.}}\quad \quad
    &g_{i}\left(\mathbf{y},\mathbf{H}\right)\leq0.
\end{align}
\end{subequations}}
\hspace{-4pt}Regarding constraint \eqref{P2c}, it prevents the transmitting power overload of each SBS.

In this work, we focus on the weighted sum rate maximization rather than the power minimization. Therefore, each SBS is supposed to perform full power transmission to maximize the transmission rate without considerations about energy conservation. Note that, full power transmission is a desired target for the total transmission power of each SBS, while the transmission power allocation for users under this target are implicitly conducted in the specific beamforming vector $\mathbf{w}$. Hence, $l_{b}\left(\mathbf{y}\right)=\sum_{n=1}^{N}\sum_{i=1}^{I}\left\Vert v_{n,i}^{b}\mathbf{w}_{n,i}^{b}\right\Vert _{2}^{2}-P_{\textrm{max}}^{b}$ is expected to approach zero. For $\forall b\in\mathcal{B}$, constraint \eqref{P2c} can be converted to a minimization learning task, denoted as
{\small \begin{subequations}\label{task3}
\begin{align}
\underset{\mathbf{y}}{\min} \quad\quad
& \left|l_{b}\left(\mathbf{y}\right)\right|,\\
    {\rm{s.t.}}\quad \quad
    &l_{b}\left(\mathbf{y}\right)\leq0.
\end{align}
\end{subequations}}
\hspace{-4pt}Note from \eqref{task1}, \eqref{task2} and \eqref{task3} that, the multi-task optimization problem consists of $I+B+1$ tasks due to the fact that there are one optimization objective and $I+B$ constraints in problem \eqref{P2}.

\vspace{-5pt}
\section{METHODOLOGY}\label{method}

In this section, we introduce the methodology for solving the multi-task learning problem. Firstly, we design two novel loss functions, i.e., NFL loss and EL loss, for the preparation of solving the multi-task problem in an end-to-end self-supervised learning manner. After that, we propose a CMTSSL algorithm to solve the underlying problem efficiently while avoiding costly manual labeling. Furthermore, we design a DMTSSL algorithm to reduce the computational complexity as well as improve the system performance. Finally, to handle the dynamic scenario, that a new SBS is suddenly added to the communication system, at a low cost, we further design a DATL algorithm by exploiting the scalability of the DMTSSL algorithm.

\subsection{Loss Design for Self-supervised Learning}\label{Lossdes}

In supervised learning, the input of regular loss functions 
is the residual between labels and network outputs. 
However, it is difficult to obtain real labels in our problem. Therefore, we need to design new loss functions to solve the problem \eqref{P2} in a self-supervised manner. The design of new loss functions should ensure that during the network training process, as the function output approaches zero, the loss function input can gradually approach the target domain without labels. To this end, we herein design two novel loss functions, i.e., NFL loss and EL loss, and introduce one classical loss function, i.e., Huber loss. Specifically, the NFL loss is defined by
\begin{sequation}\label{NFLloss}
L_{\rm{NFL}}\left(x\right)=\begin{cases}
\frac{1}{x_{1}^{2}}x-\frac{1}{x_{1}}, & x\geq x_{1}\\
-\frac{1}{x}, & x<x_{1}<0
\end{cases}.
\end{sequation}
\hspace{-4pt}The EL loss is defined by
\begin{sequation}\label{ELloss}
L_{\rm{EL}}\left(x\right)=\begin{cases}
e^{x_{2}}\left(x+1-x_{2}\right), & x\geq x_{2}\\
e^{x}, & x<x_{2}
\end{cases}.
\end{sequation}
\hspace{-4pt}The Huber loss \cite{Huber64}, which is used to assist in the construction of subsequent training loss schemes, can be expressed as
\begin{sequation}
L_{\rm{Huber}}\left(x\right)=\begin{cases}
\left|x\right|-\frac{x_{3}}{2}, & \left|x\right|\geq x_{3}>0\\
\frac{1}{2x_{3}}x^{2}, & \left|x\right|<x_{3}
\end{cases}.
\end{sequation}
\hspace{-4pt}$x_1$, $x_2$ and $x_3$ are the adjustable hyper-parameters in the NFL loss, EL loss and Huber loss, respectively. $x$ is the input of the loss function. Moreover, to support following self-supervised training loss design, some detailed analyses about the NFL and EL loss such as robustness and target domain are provided in \textbf{Appendix \ref{apend_A}}.

\subsection{Centralized Self-supervised Multi-task Learning}

\begin{figure}[!t]
\centering
\includegraphics[width=0.5\columnwidth,angle=90]{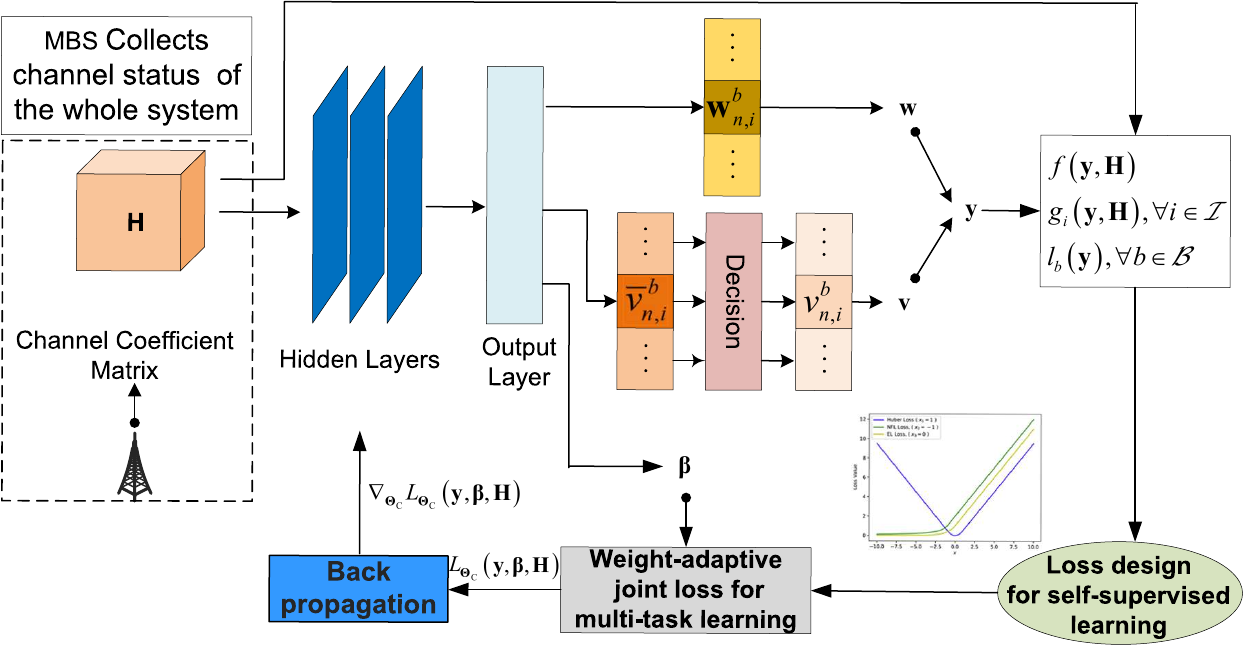}
\caption{Illustration of the CMTSSL algorithm.}
\label{CMTSSL}
\end{figure}

In this subsection, we first design training losses for the multiple tasks. Then, we develop a CMTSSL algorithm based on the NFL and EL losses to solve problem \eqref{P2} via centralized end-to-end self-supervised. The framework of the CMTSSL algorithm is illustrated in Fig. \ref{CMTSSL}.

\subsubsection{Loss schemes for the multiple tasks}
According to \eqref{task1}, \eqref{task2} and \eqref{task3}, we define the self-supervised loss for each learning task based on the newly designed losses. Specifically, based on the loss properties discussed in \textbf{Appendix \ref{apend_A}}, losses of the $I+1$ tasks in \eqref{task1} and \eqref{task2} can be defined by the NFL loss or the EL loss, while losses of the $B$ learning tasks in \eqref{task3} can be defined by the Huber loss. Thus, we give following two loss schemes for the conversion of problem \eqref{P2},

\begin{itemize}
	\item Loss Scheme 1: 

{\small \begin{subequations}\label{lossSch1}
\begin{equation}
L_{\mathbf{\Theta}}^{0}\left(\mathbf{y},\mathbf{H}\right)=L_{\textrm{NFL}}\left(f\left(\mathbf{y},\mathbf{H}\right)\right),
\end{equation}
\begin{equation}
L_{\mathbf{\Theta}}^{1,i}\left(\mathbf{y},\mathbf{H}\right)=L_{\textrm{NFL}}\left(g_{i}\left(\mathbf{y},\mathbf{H}\right)\right),\forall i\in\mathcal{I},
\end{equation}
\begin{equation}
L_{\mathbf{\Theta}}^{2,b}\left(\mathbf{y}\right)=L_{\textrm{Huber}}\left(l_{b}\left(\mathbf{y}\right)\right),\forall b\in\mathcal{B}.
\end{equation}
\end{subequations}}
	
	\item Loss Scheme 2:

{\small \begin{subequations}\label{lossSch2}
\begin{equation}
L_{\mathbf{\Theta}}^{0}\left(\mathbf{y},\mathbf{H}\right)=L_{\textrm{EL}}\left(f\left(\mathbf{y},\mathbf{H}\right)\right)
\end{equation}
\begin{equation}
L_{\mathbf{\Theta}}^{1,i}\left(\mathbf{y},\mathbf{H}\right)=L_{\textrm{EL}}\left(g_{i}\left(\mathbf{y},\mathbf{H}\right)\right),\forall i\in\mathcal{I},
\end{equation}
\begin{equation}
L_{\mathbf{\Theta}}^{2,b}\left(\mathbf{y}\right)=L_{\textrm{Huber}}\left(l_{b}\left(\mathbf{y}\right)\right),\forall b\in\mathcal{B}.
\end{equation}
\end{subequations}}
\end{itemize}

Note that equations \eqref{lossSch1} and \eqref{lossSch2} are general expressions that can be applied to all the proposed algorithms in this work. It is not difficult to observe from \eqref{task1}, \eqref{task2} and \eqref{task3} that the multi-task learning problem contains $I+B+1$ tasks, which is also equal to the total number of the objective and constraints in problem \eqref{P1}. The self-supervised loss function of each learning task can be defined by the aforesaid Loss Scheme 1 or Loss Scheme 2. 

Then, based on maximizing the Gaussian likelihood with homoscedastic uncertainty \cite{Cipo18}, the weight-adaptive joint loss for the investigated multi-task learning problem can be given by
\begin{sequation}\label{Jointloss}
\begin{aligned}
	L_{\mathbf{\Theta}}\left(\mathbf{\mathbf{y},\boldsymbol{\beta},\mathbf{H}}\right)= & \frac{1}{\left(I+B+1\right)\left(\beta_{0}\right)^{2}}L_{\mathbf{\Theta}}^{0}\left(\mathbf{y},\mathbf{H}\right)\\+&\sum_{i=1}^{I}\frac{1}{\left(I+B+1\right)\left(\beta_{1,i}\right)^{2}}L_{\mathbf{\Theta}}^{1,i}\left(\mathbf{y},\mathbf{H}\right)\\+
	 &\sum_{b=1}^{B}\frac{1}{\left(I+B+1\right)\left(\beta_{2,b}\right)^{2}}L_{\mathbf{\Theta}}^{2,b}\left(\mathbf{y}\right)\\+&\textrm{log}\left(\beta_{0}\cdot\prod_{i=1}^{I}\beta_{1,i}\prod_{b=1}^{B}\beta_{2,b}\right),
\end{aligned}
\end{sequation}
\hspace{-4pt}where $\boldsymbol{\beta}=\left[\beta_{0},\beta_{1,1},\cdots\beta_{1,I},\beta_{2,1},\cdots\beta_{2,B}\right]^{\rm{T}}\in\mathbb{R^{+}}^{\left(I+B+1\right)\times1}$ is a learnable scale vector which can be predicted by the NN together with $\mathbf{y}$. The detailed derivation of equation \eqref{Jointloss} can be found in \textbf{Appendix \ref{apend_B}}. Note from the mathematical derivation, $\boldsymbol{\beta}$ is actually the observation noise parameter vector of the network outputs, while from the expression of equation \eqref{Jointloss}, we can interpret $\boldsymbol{\beta}$ as the relative weight vector of multiple losses. When we train the NN using the joint loss function \eqref{Jointloss}, the learnable parameters $\mathbf{\Theta}$ are updated based on the gradient descent method while the vector $\boldsymbol{\beta}$ is also adaptively updated to yield a stable convergence. The loss function \eqref{Jointloss} will converge towards zero due to the components of the loss function \eqref{Jointloss} being all nonnegative, which guarantees stable training.

\subsubsection{Training process of the CMTSSL algorithm}
\begin{algorithm*}[!t]
{\small
\caption{The training process of the CMTSSL algorithm}
\label{alg_LDB_CSSMTL}
\begin{algorithmic}[1]
 \STATE  \textbf{Initialization:} Randomly initialize the NN parameters to $\mathbf{\Theta}^{\left(0\right)}_{\textrm{C}}$. Let $\mathbf{\Theta}=\mathbf{\Theta}^{\left(0\right)}_{\textrm{C}}$.
  \STATE  \textbf{Input:} Channel realization dataset $\mathcal{D}_{\textrm{C}}$.
 \STATE \textbf{For} training epoch $ t = 1, 2, \ldots,\varGamma $ \textbf{do:}
 \STATE \quad \quad \textbf{For} training step $\ell = 0,1,\cdots,\frac{\left|\mathcal{D}_{\textrm{C}}\right|}{S}$ \textbf{do:} 
\STATE \quad \quad \quad \quad Randomly extract mini-batch samples $\mathcal{D}_{\textrm{C}}^{\textrm{mb}}$ for training.  
\STATE \quad \quad \quad \quad Obtain $S$ outputs $\{\left.\mathbf{y}^{\left(s\right)},\boldsymbol{\mathbf{\beta}}^{\left(s\right)}\right|s\in\mathcal{S}\}$ on each input sample $\mathbf{H}^{\left(s\right)}\in\mathcal{D}_{\textrm{C}}^{\textrm{mb}}$ by \eqref{output}.
\STATE \quad \quad \quad \quad Compute the target values in mini-batch by \eqref{P2}
{\small \[\quad \quad \quad \quad \left\{ \left.f\left(\mathbf{y}^{\left(s\right)},\mathbf{H}^{\left(s\right)}\right),\left\{ \left.g_{i}\left(\mathbf{y}^{\left(s\right)},\mathbf{H}^{\left(s\right)}\right)\right|i\in\mathcal{I}\right\} ,\left\{ \left.l_{b}\left(\mathbf{y}^{\left(s\right)}\right)\right|b\in\mathcal{B}\right\} \right|s\in\mathcal{S}\right\}.\]}
\STATE \quad \quad \quad \quad Compute the multi-task losses in mini-batch by \eqref{lossSch1} \textbf{or} \eqref{lossSch2}
{\small \[\quad \quad \quad \quad\left\{ \left.L_{\mathbf{\Theta}_{\textrm{C}}^{\left(l\right)}}^{0},\left\{ \left.L_{\mathbf{\Theta}_{\textrm{C}}^{\left(l\right)}}^{1,i}\right|i\in\mathcal{I}\right\} ,\left\{ \left.L_{\mathbf{\Theta}_{\textrm{C}}^{\left(l\right)}}^{2,b}\right|b\in\mathcal{B}\right\} \right|s\in\mathcal{S}\right\} .\]}
\STATE \quad \quad \quad \quad Calculate $L^{\textrm{avg}}\left(\mathbf{\Theta}_{\textrm{C}}^{\left(\ell\right)}\right)$ the mini-batch average of the joint losses by \eqref{Jointloss} and \eqref{avgjointloss}.
\STATE \quad \quad \quad \quad Update ${\mathbf{\Theta}_{\textrm{C}}}$ by $\nabla_{\mathbf{\Theta}_{\textrm{C}}^{\left(\ell\right)}}L^{\textrm{avg}}\left(\mathbf{\Theta}_{\textrm{C}}^{\left(\ell\right)}\right)$
{\small \[\quad \quad \quad \quad\mathbf{\Theta}_{\textrm{C}}^{\left(\ell+1\right)}\leftarrow\mathbf{\Theta}_{\textrm{C}}^{\left(\ell\right)}-\gamma_{\textrm{C}}\nabla_{\mathbf{\Theta}_{\textrm{C}}^{\left(\ell\right)}}L^{\textrm{avg}}\left(\mathbf{\Theta}_{\textrm{C}}^{\left(\ell\right)}\right)\]}
\STATE \quad \quad \textbf{End For}
\STATE \textbf{End For}
\end{algorithmic}}
\end{algorithm*}
The CMTSSL algorithm is deployed on the MBS to perform the centralized offline training and online test. The dataset for training the CMTSSL algorithm is a compact set of channel realizations, which is denoted as $\mathcal{D}_{\textrm{C}}=\left\{ \left.\mathbf{H}^{\left(k\right)}\right|k\in\mathbb{Z}_{0+}\right\}$. $\mathbf{H}^{\left(k\right)}$ is the channel coefficients of the communication system at the $k$-th realization. Firstly, the MBS collects the channel coefficients $\mathbf{H}$ of the entire communication system and feeds it to the NN. Consequently, the output layer produces three elements, i.e., $\mathbf{w}$, $\mathbf{v}$, and $\boldsymbol{\beta}$. The output of the NN in the CMTSSL algorithm can be expressed as
\begin{sequation}\label{output}
\left\{ \mathbf{y},\boldsymbol{\beta}\right\} =\pi_{\mathbf{\Theta}_{\textrm{C}}}\left(\mathbf{H}\right),
\end{sequation}
\hspace{-4pt}where $\pi_{\mathbf{\Theta}_{\textrm{C}}}\left(\cdot\right)$ denotes the parameterized NN and $\mathbf{\Theta}_{\textrm{C}}$ denotes the learnable parameter of the NN in the CMTSSL framework. Note that the output of the NN is a continuous-valued vector in practice while the SA indicator $v_{n,i}^{b}$ is required to be a binary value of 0/1. Hence, for the SA part in the output elements, we add an additional decision layer behind the output layer to convert $\overline{v}_{n,i}^{b}$ to the binary values, i.e., $v_{n,i}^{b}$. Subsequently, we can obtain the current target value of each task, i.e., $f\left(\mathbf{y},\mathbf{H}\right)$, $g_{i}\left(\mathbf{y},\mathbf{H}\right),\forall i\in\mathcal{I}$ and $l_{b}\left(\mathbf{y}\right),\forall b\in\mathcal{B}$ in the multi-task learning problems \eqref{task1}, \eqref{task2} and \eqref{task3}, based on the input $\mathbf{H}$ and the output $\mathbf{y}$ as well as the formulations in problem \eqref{P2}. Then, let $\mathbf{\Theta}=\mathbf{\Theta}_{\textrm{C}}$, the loss values of each learning task can be computed according to \eqref{lossSch1} in the \textbf{Loss Scheme 1} or \eqref{lossSch2} in the \textbf{Loss Scheme 2}. Finally, the weight-adaptive joint loss of the multi-task learning problem can be obtained by \eqref{Jointloss}, where another component of the NN output, i.e., $\boldsymbol{\beta}$, contributes to the loss weighs in the joint loss.

During the practical training, the MBS randomly extracts $S$ samples from $\mathcal{D}_{\textrm{C}}$ as a mini-batch for each training iteration. The mini-batch average of the joint loss \eqref{Jointloss} is adopted to yield a more stable convergence. That is, for any mini-batch set $\mathcal{D}_{\textrm{C}}^{\textrm{mb}}=\left\{ \left.\mathbf{H}^{\left(s\right)}\right|s\in\mathcal{S}\subseteq\mathbb{Z}_{0+},\left|\mathcal{S}\right|=S\right\} $, we have
\begin{sequation}\label{avgjointloss}
L^{\textrm{avg}}\left(\mathbf{\Theta}_{\textrm{C}}\right)=\frac{1}{S}\sum_{s\in\mathcal{D}_{\textrm{C}}^{\textrm{mb}}}L_{\mathbf{\Theta}_{\textrm{C}}}\left(\mathbf{y}^{\left(s\right)},\boldsymbol{\mathbf{\beta}}^{\left(s\right)},\mathbf{H}^{\left(s\right)}\right),
\end{sequation}
\hspace{-4pt}where $\left\{ \mathbf{y}^{\left(s\right)},\boldsymbol{\mathbf{\beta}}^{\left(s\right)}\right\}$ is the output when the NN is fed with the $s$-th sample in $\mathcal{D}_{\textrm{C}}^{\textrm{mb}}$. Thus, we optimize $\mathbf{\Theta}_{\textrm{C}}$ by minimizing this average loss and $\mathbf{\Theta}_{\textrm{C}}$ can be updated by $\nabla_{\mathbf{\Theta}_{\textrm{C}}}L^{\textrm{avg}}\left(\mathbf{\Theta}_{\textrm{C}}\right)$ in the back propagation (BP) process. The overall process of the proposed CMTSSL algorithm is summarized in \textbf{Algorithm \ref{alg_LDB_CSSMTL}}. $\varGamma$ is the number of training epochs which depends on the size of $\mathcal{D}_{\textrm{C}}$ and $\mathcal{D}_{\textrm{C}}^{\textrm{mb}}$. Let $\left|\mathcal{D}_{\textrm{C}}\right|$ denote the size of set $\mathcal{D}_{\textrm{C}}$. $\gamma_{\textrm{C}}$ is the learning rate of the CMTSSL algorithm and $\mathbf{\Theta}_{\textrm{C}}^{\left(\ell\right)}$ denotes the NN parameters in the $\ell$-th iteration. It is worth mentioning that the newly designed NFL/EL losses as well as the proposed CMTSSL algorithm are not merely applicable for solving the specific problem \eqref{P1} but also general for any other problems with the general form as the formulation \eqref{P2}.

\subsection{Distributed Self-supervised Multi-task Learning}
\begin{figure}[!t]
\centering
\includegraphics[width=0.5\columnwidth,angle=90]{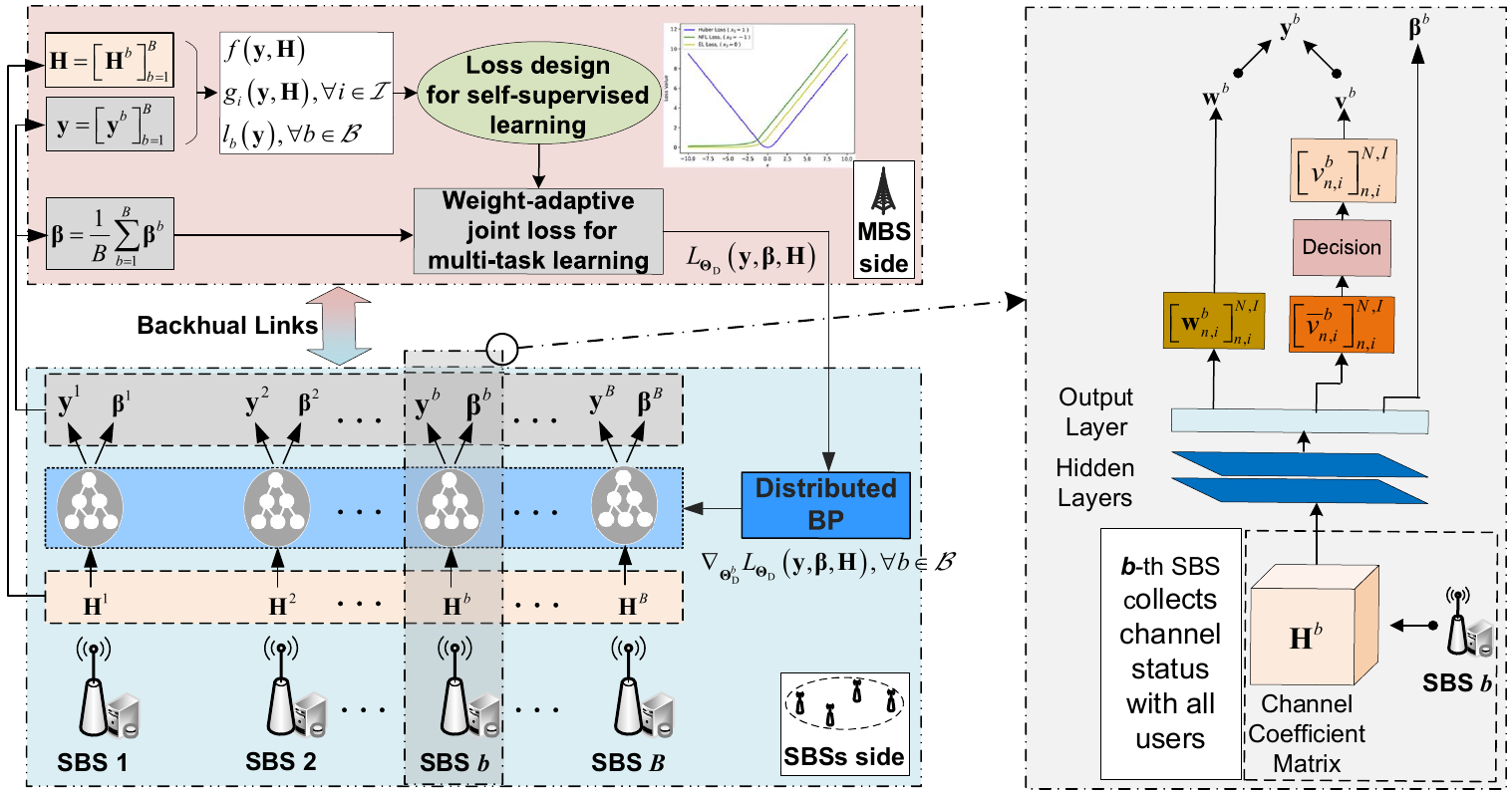}
\caption{Illustration of the DMTSSL algorithm.}
\label{DMTSSL}
\end{figure}

Inspired by a classic distributed algorithm in the multi-agent reinforcement learning, namely centralized training and decentralized execution \cite{LoweR17}, we herein design a distributed DL framework, i.e., DMTSSL algorithm as illustrated in Fig. \ref{DMTSSL}, where the SBSs and the MBS play the similar roles of the distributed actors and centralized critic, respectively. In the DMTSSL algorithm, each SBS can individually determine its own joint SA and BF vector by deploying an NN model with the same structure, while the MBS computes the global loss to supervise the training of each local NN model during the phase of offline training. The MEC servers with sufficient computational resources provide supporting infrastructure for the design of the DMTSSL algorithm.

Specifically, let $\mathbf{\Theta}_{\textrm{D}}=\left\{ \mathbf{\Theta}_{\textrm{D}}^{b}\right\} _{b=1}^{B}$ denote the stack of the learnable parameters set of the NN at each SBS. Each SBS $b\in\mathcal{B}$ collects the channel status with all users within the service area independently. Then the NN deployed at each SBS $b$ is fed with the collected channel coefficient matrix $\mathbf{H}^{b}$ and consequently, outputs the joint SA and BF vector of the SBS $b$, i.e., $\mathbf{y}^{b}=\left\{ \mathbf{w}^{b},\mathbf{v}^{b}\right\}$, as well as the loss weights $\boldsymbol{\beta}^{b}=\left[\beta_{0}^{b},\beta_{1,1}^{b},\cdots\beta_{1,I}^{b},\beta_{2,1}^{b},\cdots\beta_{2,B}^{b}\right]$, denoted as
\begin{sequation}\label{D_output}
\left\{ \mathbf{y}^{b},\boldsymbol{\beta}^{b}\right\} =\pi_{\mathbf{\Theta}_{\textrm{D}}^{b}}\left(\mathbf{H}^{b}\right).
\end{sequation}
\hspace{-4.5pt}After that, for each SBS $b\in\mathcal{B}$, the $\mathbf{H}^{b}$, $\mathbf{y}^{b}$ and $\boldsymbol{\mathbf{\beta}}^{b}$ are uploaded to the MBS via backhaul links for loss calculation. At the MBS, the received $\mathbf{H}^{b}, \forall b\in\mathcal{B}$ and $\mathbf{y}^{b},\forall b\in\mathcal{B}$ are first stacked to $\mathbf{H}$ and $\mathbf{y}$, respectively. Then, on the basis of the $\mathbf{H}$ and $\mathbf{y}$, the current target value of each learning task, i,e., $f\left(\mathbf{y},\mathbf{H}\right),g_{i}\left(\mathbf{y},\mathbf{H}\right),l_{b}\left(\mathbf{y}\right),\forall i\in\mathcal{I},\forall b\in\mathcal{B}$, can be obtained. Subsequently, when we set $\mathbf{\Theta}=\mathbf{\Theta}_{\textrm{D}}$, the loss values in the \textbf{Loss Scheme 1} or in the \textbf{Loss Scheme 2} can be computed by \eqref{lossSch1} or \eqref{lossSch2} respectively. Likewise, we aggregate the received $\left\{ \boldsymbol{\mathbf{\beta}}^{b}\right\} _{b=1}^{B}$ to the final weights $\boldsymbol{\beta}$ by
\begin{sequation} \label{weightedLossweights}
\boldsymbol{\mathbf{\beta}}=\frac{1}{B}\sum_{b=1}^{B}\omega^{b}\boldsymbol{\mathbf{\beta}}^{b}.
\end{sequation}
\hspace{-4.5pt}In this work, we assume that there is no priority among SBSs. Hence, we reasonably set $\omega^{b}=1,\forall b\in\mathcal{B}$. Finally, the weight-adaptive joint loss of the multi-task learning problem can be obtained by \eqref{Jointloss}. Once the joint loss is obtained, the MBS broadcast $L_{\mathbf{\Theta}_{\textrm{D}}}\left(\mathbf{\mathbf{y},\mathbf{H},\mathbf{\beta}}\right)$ back to all the SBSs via the backhual link. Then, each SBS updates its NN parameters $\mathbf{\Theta}_{\textrm{D}}^{b}$ in parallel based on the distributed gradient calculation, i.e., $\nabla_{\mathbf{\Theta}_{\textrm{D}}^{b}}L_{\mathbf{\Theta}_{\textrm{D}}}\left(\mathbf{\mathbf{y},\mathbf{H},\mathbf{\beta}}\right),\forall b\in\mathcal{B}$.

\begin{algorithm*}[!t]
{\small
\caption{The training process of the DMTSSL algorithm}
\label{alg_LDB_DSSMTL}
\begin{algorithmic}[1]
\STATE  \textbf{Initialization:} Randomly initialize the NN parameters at each SBS to $\mathbf{\Theta}_{\textrm{D}}^{b,{\left(0\right)}}$. Let $\mathbf{\Theta}=\mathbf{\Theta}^{\left(0\right)}_{\textrm{D}}$.
\STATE  \textbf{Distributed Input:} Channel realization dataset at each SBS, $\mathcal{D}_{\textrm{D}}^{b},\forall b\in\mathcal{B}$.
\STATE \textbf{For} training epoch $ t = 1, 2, \ldots,\varGamma $ \textbf{do:}
\STATE \quad \quad \textbf{For} training step $\ell = 0,1,\cdots,\frac{\left|\mathcal{D}_{\textrm{D}}^{b,\textrm{mb}}\right|}{S}$ \textbf{do:}
\STATE \quad \quad \quad \quad The MBS \textbf{do:}
\STATE \quad \quad \quad \quad \quad \quad \textbf{If} receive $\left\{ \left.\mathbf{y}^{b,\left(s\right)},\boldsymbol{\mathbf{\beta}}^{b,\left(s\right)},\mathbf{H}^{b,\left(s\right)}\right|\mathbf{H}^{b,\left(s\right)}\in\mathcal{D}_{\textrm{D}}^{b,\textrm{mb}},s\in\mathcal{S}\right\} $ from SBSs \textbf{then}
\STATE \quad \quad \quad \quad \quad \quad \quad \quad For $\forall s\in\mathcal{S}$, concatenate $\mathbf{y}^{b,\left(s\right)},\mathbf{H}^{b,\left(s\right)},\forall b\in\mathcal{B}$ into $\mathbf{y}^{\left(s\right)},\mathbf{H}^{\left(s\right)}$ and obtain $\boldsymbol{\mathbf{\beta}}^{\left(s\right)}$ by \eqref{weightedLossweights}.
\STATE \quad \quad \quad \quad \quad \quad \quad \quad Compute mini-batch target values by \eqref{P2}.
\STATE \quad \quad \quad \quad \quad \quad \quad \quad Compute the mini-batch multi-task losses by \eqref{lossSch1} \textbf{or} \eqref{lossSch2}.
\STATE \quad \quad \quad \quad \quad \quad \quad \quad Obtain mini-batch joint losses $\left\{ \left.L_{\mathbf{\Theta}_{\textrm{D}}^{\left(\ell\right)}}\left(\mathbf{y}^{\left(s\right)},\boldsymbol{\mathbf{\beta}}^{\left(s\right)},\mathbf{H}^{\left(s\right)}\right)\right|s\in\mathcal{S}\right\}$ by \eqref{Jointloss}.
\STATE \quad \quad \quad \quad \quad \quad \quad \quad Obtain $L^{\textrm{avg}}\left(\mathbf{\Theta}_{\textrm{D}}^{\left(\ell\right)}\right)$ by \eqref{avg_D_jointloss} and then broadcast it to all SBSs.
\STATE \quad \quad \quad \quad \quad \quad \textbf{End If}
 \STATE \quad \quad \quad \quad The MEC server at each SBS $b\in\mathcal{B}$ \textbf{in parallel do:}
\STATE \quad \quad \quad \quad \quad \quad \textbf{If} receive $L^{\textrm{avg}}\left(\mathbf{\Theta}_{\textrm{D}}^{\left(\ell\right)}\right)$ broadcasted from the MBS \textbf{then}
\STATE \quad \quad \quad \quad \quad \quad \quad \quad Update its local NN parameters by
{\small \[\mathbf{\Theta}_{\textrm{D}}^{b,\left(\ell+1\right)}\leftarrow\mathbf{\Theta}_{\textrm{D}}^{b,\left(\ell\right)}-\gamma_{\textrm{D}}\nabla_{\mathbf{\Theta}_{\textrm{D}}^{b,\left(\ell\right)}}L^{\textrm{avg}}\left(\mathbf{\Theta}_{\textrm{D}}^{\left(\ell\right)}\right)\]}
\STATE \quad \quad \quad \quad \quad \quad \quad \quad Randomly extract mini-batch samples $\mathcal{D}_{\textrm{D}}^{b,\textrm{mb}}$ for training. 
\STATE \quad \quad \quad \quad \quad \quad \quad \quad Output $\left\{ \left.\mathbf{y}^{b,\left(s\right)},\boldsymbol{\mathbf{\beta}}^{b,\left(s\right)}\right|\mathbf{H}^{b,\left(s\right)}\in\mathcal{D}_{\textrm{D}}^{b,\textrm{mb}},s\in\mathcal{S}\right\}$ in mini-batch by \eqref{D_output}. 
\STATE \quad \quad \quad \quad \quad \quad \quad \quad Upload the mini-batch inputs and outputs to the MBS.
\STATE \quad \quad \quad \quad \quad \quad \textbf{End If}
\STATE \quad \quad \textbf{End For}
\STATE \textbf{End For}
\end{algorithmic}}
\end{algorithm*}

Moreover, the mini-batch training is also adopted in this DMTSSL algorithm. Concretely, in the framework of the DMTSSL algorithm, each SBS $b\in\mathcal{B}$ holds a training dataset, denoted as $\mathcal{D}_{\textrm{D}}^{b}=\left\{ \left.\mathbf{H}^{b,\left(k\right)}\right|k\in\mathbb{Z}_{0+}\right\}$. For one training iteration, each SBS $b$ randomly extracts a mini-batch samples $\mathcal{D}_{\textrm{D}}^{b,\textrm{mb}}=\left\{ \left.\mathbf{H}^{b,\left(s\right)}\right|s\in\mathcal{S}\subseteq\mathbb{Z}_{0+},\left|\mathcal{S}\right|=S\right\} $ from its local dataset. Then, let $\mathbf{\Theta}=\mathbf{\Theta}_{\textrm{D}}$ in \eqref{Jointloss}, the average joint loss of a minibatch can be expressed as
\begin{sequation} \label{avg_D_jointloss}
L^{\textrm{avg}}\left(\mathbf{\Theta}_{\textrm{D}}\right)=\frac{1}{S}\sum_{s=1}^{S}L_{\mathbf{\Theta}_{\textrm{D}}}\left(\mathbf{y}^{\left(s\right)},\boldsymbol{\mathbf{\beta}}^{\left(s\right)},\mathbf{H}^{\left(s\right)}\right).
\end{sequation}
\hspace{-4pt}Then, $\mathbf{\Theta}_{\textrm{D}}$ can be optimized by minimizing the $L^{\textrm{avg}}\left(\mathbf{\Theta}_{\textrm{D}}\right)$ and the learnable parameters $\mathbf{\Theta}_{\textrm{D}}^{b}$ at each SBS $b$ can be parallelly updated by $\nabla_{\mathbf{\Theta}_{\textrm{D}}^{b}}L^{\textrm{avg}}\left(\mathbf{\Theta}_{\textrm{D}}\right)$ during the distributed BP process. The overall process of the proposed DMTSSL algorithm is summarized in \textbf{Algorithm \ref{alg_LDB_DSSMTL}}, where $\left|\mathcal{D}_{\textrm{D}}^{b,\textrm{mb}}\right|$ denotes the size of set $\mathcal{D}_{\textrm{D}}^{b,\textrm{mb}}$ and $\mathbf{\Theta}_{\textrm{D}}^{\left(b,\ell\right)}$ are the NN parameters at the SBS $b$ in the $\ell$-th iteration. $\gamma_{\textrm{D}}$ is the learning rate of the DMTSSL algorithm.

\textbf{Comparisons between the CMTSSL and the DMTSSL:}
From the framework of the DMTSSL algorithm, we can find that the input and output of the NN at each SBS $b$ are $\mathbf{H}^{b}$ and $\mathbf{y}^{b}$ whose dimensions are only $\frac{1}{B}$ of those in the CMTSSL algorithm. Consequently, the dimensions of the input and output layers in the DMTSSL algorithm are $\frac{1}{B}$ of those in the CMTSSL algorithm, which contributes to that the model complexity in the DMTSSL algorithm is significantly lower than that in the CMTSSL algorithm. Concurrently, the MEC servers on SBSs can support the distributed learning so as not to increase the time complexity of the DMTSSL algorithm. Moreover, the framework design in the proposed DMTSSL algorithm can further improve the system performance. Detailedly, on the one hand, the NN model on each SBS can obtain the loss with global information to ensure the parameters training towards the direction of maximizing the sum rate of the whole system rather than this single SBS. On the other hand, the NN model on each SBS only needs to extract features from the local channel $\mathbf{H}^{b}$ and output the joint SA and BF policy of this single SBS $\mathbf{y}^{b}$. Compared with global input $\mathbf{H}$ and global output $\mathbf{y}$ in the proposed CMRSSL algorithm, this distributed framework in the proposed DMRSSL algorithm can split the global decision-making task into many joint local decision-making tasks. Then, the DMRSSL algorithm accurately disperses these local tasks into many distributed local models so as to significantly reduce the training and fitting difficulties of the NN networks, and thus achieves remarkable performance gain. Thirdly, the distributed dataset $\left\{ \mathcal{D}_{\textrm{C}}^{b}\right\} _{b=1}^{B}$ for the DMRSSL algorithm is derived from the mixed dataset $\mathcal{D}_{\textrm{C}}$ based on the spatial locations of all the SBSs instead of a random division of dataset $\mathcal{D}_{\textrm{C}}$. This data preprocessing with {\em a priori} knowledge of the physical world can help reduce the difficulty of feature extraction and bring extra performance gain \cite{Peng21}. Finally, the distributed framework also provides high scalability for the DMTSSL algorithm, which will be verified and exploited in the following DATL algorithm.

\subsection{Distance-aware Transfer Learning}

\begin{figure}[!t]
\centering
\includegraphics[width=0.42\columnwidth,angle=90]{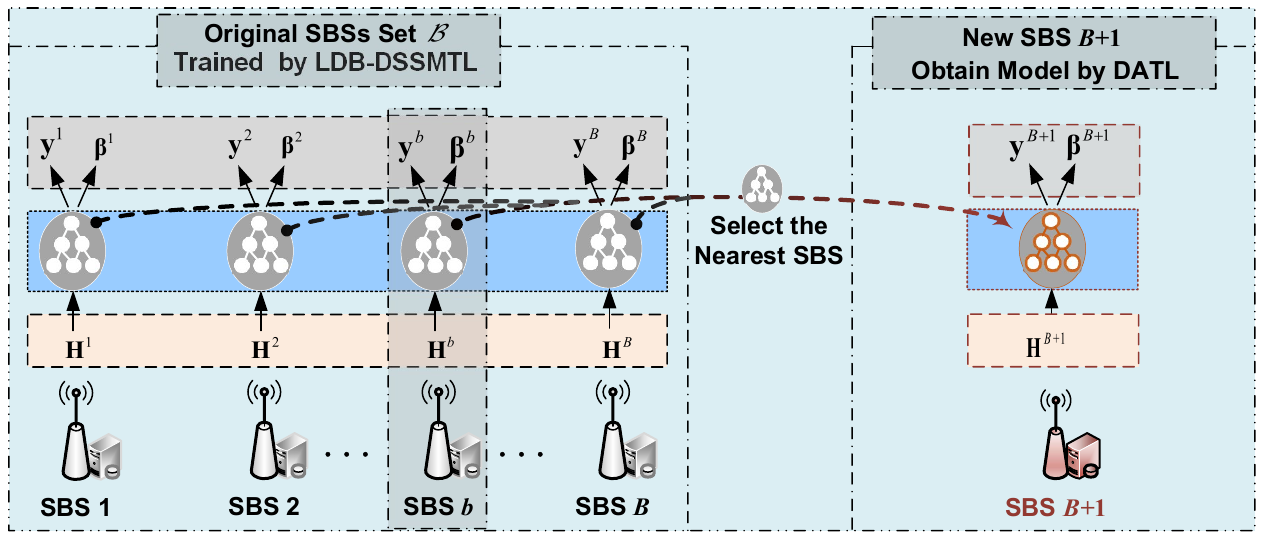}
\vspace{-1em}
\caption{Illustration of the DATL algorithm.}
\label{DATL}
\end{figure}

In practical production and life, there are some common dynamic communication scenarios where new SBSs appear in the original communication network, such as emergency communication scenarios where mobile SBSs are temporarily deployed for concerts or other large-scale events, unmanned aerial vehicle (UAV)-assisted communication scenarios where mobile UAV base station cooperates with ground SBSs for dynamic networking, and etc.. When considering these dynamic scenarios that a new SBS is added to the communication system, most learning methods need to retrain the NN model for better performance. However, the retraining process is costly in terms of time and computation. To handle this dynamic scenario with negligible cost, we develop a simple but effective DATL algorithm by exploiting the scalability of the proposed DMTSSL algorithm. The framework of the DATL algorithm is illustrated in Fig. \ref{DATL}.

Specifically, we assume that there is an original SBS set $\mathcal{B}$ where each SBS is deployed with an NN model trained by the DMTSSL algorithm. When a new SBS, assumed to place at $\left(\overline{x}_{0},\overline{y}_{0}\right)$, is added to the original SBSs set, the new SBS first senses the geographical distance from the other SBSs in the original set $\mathcal{B}$ and then transfer the NN model on the nearest SBS to itself. After that, the transferred model is directly used by the new SBS for its SA and BF. The nearest SBS can be obtained by
\begin{sequation} \label{DA}
b^{*}=\textrm{arg}\quad\underset{b\in\mathcal{B}}{\textrm{min}}\sqrt{\left|\overline{x}_{b}-\overline{x}_{0}\right|^{2}+\left|\overline{y}_{b}-\overline{y}_{0}\right|^{2}},
\end{sequation}
\hspace{-4pt}where $\left(\overline{x}_{b},\overline{y}_{b}\right)$ is the geographical coordinate of the SBS $b$. When the transfer process is completed, these $B+1$ SBSs can perform their own joint SA and BF to maximize the weight sum rate of this new system. Note that the reasonable assumption that the new SBS is the same as the original SBS on hardware configurations, i.e., the number of subcarriers and transmitting antenna, is needed for the DATL algorithm. The DATL algorithm is summarized in \textbf{Algorithm \ref{alg_DATL}}. It is not difficult to find that the whole process of the DATL algorithm is extremely simple and the extra calculation comes from the distance-aware step which consumes negligible computing resources. Moreover, the proposed DATL algorithm is remarkable in terms of the system sum-rate performance, which will be verified by simulations in the next section.

\begin{algorithm}[!t]
{\small
\caption{The process of the DATL algorithm}
\label{alg_DATL}
\begin{algorithmic}[1]
\STATE  \textbf{Initialization:} Obtain the optimal $\mathbf{\Theta}_{\textrm{D}}^{*}=\left\{ \mathbf{\Theta}_{\textrm{D}}^{b,*}\right\} _{b=1}^{B}$ on the original SBSs set $\mathcal{B}$ by the DMTSSL algorithm.
\STATE Sense the geographical locations of all SBSs, i.e., $\left(x_{0},y_{0}\right)$ and $\left(x_{b},y_{b}\right),\forall b\in\mathcal{B}$.
\STATE Select the nearest SBS by \eqref{DA}.
\STATE Model transfer $\mathbf{\Theta}_{\textrm{D}}^{0}\leftarrow\mathbf{\Theta}_{\textrm{D}}^{b^{*},*}$.
\end{algorithmic}}
\end{algorithm}

\section{NUMERICAL SIMULATIONS AND ANALYSES} \label{sec4}

In this section, we evaluate the performance of the proposed algorithms via numerical examples. We consider the $3$rd generation partnership project 38.901 urban-macrocell (3GPP 38.901 UMa) scenario to cope with realistic scenarios while the quasi-deterministic radio channel generator (QuaDRiGa) \cite{Jaeckel14,Jaeckel21} is adopted to model the channel of the wireless communication system. The simulation environment contains a single MBS, $B$ multi-antennas SBS and $I$ multi-antennas users. Each SBS has $N$ subcarriers for its download transmitting. For simplicity, we assume that all users on all subcarriers have the same noise variance, the maximum transmitting power of each SBS as well as the minimum data rate of each user are equal, i.e., $\sigma_{n,i}^{2}=\sigma^{2},\forall i\in\mathcal{I},\forall n\in\mathcal{N}$,$P_{\textrm{max}}^{b}=P_{\textrm{max}},\forall b\in\mathcal{B}$ and $r_{i}^{\textrm{min}}=r^{\textrm{min}},\forall i\in\mathcal{I}$. In addition, we assume that there is no priority among users. Hence, we reasonably set $\alpha_{i}=1,\forall i\in\mathcal{I}$. Simulation parameters are listed in detail in Table \ref{simu_setup}. 

\begin{table*}[!t]
\centering
\caption{Simulation Parameters} \label{simu_setup}
\vspace{-1em}
\begin{tabular}{c|c||c|c}
\toprule[1pt]
\textbf{Parameter}               & \textbf{Value}    & \textbf{Parameter}                                & \textbf{Value} \\ \hline\hline
Scenario                                    &  3GPP 38.901 UMa  & Channel model              &   QuaDRiGa    \\ \hline
Height of the SBS antenna                   & $25$ m            & Height of the user antenna &  $1.5$ m     \\ \hline
Center frequency                            &  $2.1$ GHz        & Baseband bandwidth         &  $20$ MHz     \\ \hline
Number of the SBS antenna $M_t$             &  4 (2V2H)         & Number of the user antenna $M_r$ &  $2$    \\ \hline
\tabincell{c}{Maximum transmitting \\power of each SBS $P_{\textrm{max}}$}    &  $40$ dBm                   & \tabincell{c}{Minimum data rate \\of each user $r^{\textrm{min}}$}                     &  $0.02$ bps/Hz     \\ \hline
Noise power $\sigma^{2}$                    &   $-26$ dBm       & Batch size  $S$            &  $100$    \\ \hline
Learning rate $\gamma_{\textrm{C}}$, $\gamma_{\textrm{D}}$                    & $10^{-2}$, $10^{-3}$, resp. & Loss hyper-parameters $x_1,x_2,x_3$                                                  &  $-1,0,0.11$, resp.     \\ \hline
\bottomrule[1pt]
\end{tabular}
\vspace{-1.5em}
\end{table*}

\subsection{Network Configurations}

The dense NN with three hidden layers is adopted in the proposed CMTSSL algorithm, where each hidden layer contains a fully connected (FC) layer, a batch normalization layer and the {\em ReLU} active function. The three successive FC layers are set with 512, 1024, and 512 neural cells, respectively. As for the output layer of the deep dense NN in the proposed CMTSSL algorithm, we adopt the FC layer with the {\em Sigmoid} active function. Likewise, the same hidden layer architecture is also adopted in each local NN in the distributed DMTSSL algorithm. Note that, for the proposed CMTSSL algorithm and the proposed CMTSSL algorithm, the sizes of the input layer and output layer depend on the sizes of their input and output data. Adam optimizer \cite{King15} is used to update the parameters $\mathbf{\Theta}_{\textrm{c}}$ and $\mathbf{\Theta}_{\textrm{D}}$.


\subsection{Baselines}
\subsubsection{Loss Baselines}
To verify the superiority of the proposed self-supervised Loss Scheme 1 and Loss Scheme 2, we set following three loss baselines for comparisons:
\begin{itemize}
	\item Loss Baseline 1: The $f\left(\mathbf{y},\mathbf{H}\right)$, $g_{i}\left(\mathbf{y},\mathbf{H}\right)$ and $l_{b}\left(\mathbf{y}\right)$ are directly fed to the joint loss \eqref{Jointloss} without any self-supervised losses conversion, defined as:
\begin{sequation}\label{lossBas1}
\begin{aligned}
L_{\mathbf{\Theta}}^{0}\left(\mathbf{y},\mathbf{H}\right)=& f\left(\mathbf{y},\mathbf{H}\right) \\
L_{\mathbf{\Theta}}^{1,i}\left(\mathbf{y},\mathbf{H}\right)=& g_{i}\left(\mathbf{y},\mathbf{H}\right),\forall i\in\mathcal{I}\\
L_{\mathbf{\Theta}}^{2,b}\left(\mathbf{y}\right)=& l_{b}\left(\mathbf{y}\right),\forall b\in\mathcal{B}
\end{aligned}
\end{sequation}
	
	\item Loss Baseline 2: Considering that the NFL loss is proposed based on the negative fraction function, thereby we replace the NFL function in \eqref{lossSch1} with the negative fraction function and define another loss baseline as follows:
\begin{sequation}\label{lossBas2}
\begin{aligned}
L_{\mathbf{\Theta}}^{0}\left(\mathbf{y},\mathbf{H}\right)=&\frac{-1}{f\left(\mathbf{y},\mathbf{H}\right)}\\
L_{\mathbf{\Theta}}^{1,i}\left(\mathbf{y},\mathbf{H}\right)=&\frac{-1}{g_{i}\left(\mathbf{y},\mathbf{H}\right)},\forall i\in\mathcal{I}\\
L_{\mathbf{\Theta}}^{2,b}\left(\mathbf{y}\right)=& L_{\rm Huber}\left({l_{b}\left(\mathbf{y}\right)}\right),\forall b\in\mathcal{B}.
\end{aligned}
\end{sequation}

	\item Loss Baseline 3: Considering that the EL loss is proposed based on the exponential function, thereby we define another loss baseline by replacing the EL function in \eqref{lossSch2} with the exponential function, which can be given by:
\begin{sequation}\label{lossBas3}
\begin{aligned}
L_{\mathbf{\Theta}}^{0}\left(\mathbf{y},\mathbf{H}\right)=& e^{f\left(\mathbf{y},\mathbf{H}\right)}\\
L_{\mathbf{\Theta}}^{1,i}\left(\mathbf{y},\mathbf{H}\right)=& e^{g_{i}\left(\mathbf{y},\mathbf{H}\right)},\forall i\in\mathcal{I}\\
L_{\mathbf{\Theta}}^{2,b}\left(\mathbf{y}\right)=& L_{\rm Huber}\left({l_{b}\left(\mathbf{y}\right)}\right),\forall b\in\mathcal{B}.
\end{aligned}
\end{sequation}

\end{itemize}

\begin{table*}[!htbp]
\centering
\caption{Performance evaluation of the proposed NFL and EL losses. ($B=3,I=10,N=4$)} \label{Loss_eva}
\vspace{-1em}
\begin{threeparttable}
\begin{tabular}{@{}cccccc@{}}
\toprule[1pt]
\multirow{2}{*}{\textbf{Algorithms}} & \multicolumn{5}{c}{\textbf{Sum Rate on Different Loss Schemes (bps/Hz)}}                                                         \\ \cmidrule(l){2-6} 
                                     & \textbf{Loss Scheme 1} & \textbf{Loss Scheme 2} & Loss Baseline 1 & Loss Baseline 2 & Loss Baseline 3 \\ \midrule \midrule
CMTSSL                  & 0.796                  & 0.687                  & -- \tnote{1}             & --                       & 0.249                    \\ \midrule
DMTSSL                  & 0.911                  & 0.986                  & --                       & --                       & 0.381                    \\ \bottomrule[1pt]
\end{tabular}
 \begin{tablenotes}
        \footnotesize
        \item[1]  ``--'' indicates that the algorithm is not convergent under the corresponding loss scheme.
 \end{tablenotes}
\end{threeparttable}
\vspace{-1em}
\end{table*}

\subsubsection{Random SA and Zero-force BF (RSA+ZFBF) Algorithm} In this baseline algorithm, the SA scheme is randomly formulated while the BF scheme performs the classical zero-force beamforming algorithm with equal power allocation. The computational complexity of the algorithm can be measured by the common metric, namely floating point operations (FLOPs). It is not difficult to analyze that the FLOPs of this baseline algorithm can be given by $FLOPs_{\left(\textrm{RSA+ZFBF}\right)}\approx 6BN\left[2IM_{\textrm{t}}M_{\textrm{r}}+\left(2M_{\textrm{t}}+IM_{\textrm{r}}\right)\left(IM_{\textrm{r}}\right)^{2}\right]$.

\subsubsection{Greedy SA and Zero Force BF (GSA+ZFBF) Algorithm} In this baseline algorithm, the SA scheme performs the greedy search policy based on the channel quality, i.e., $\left|\left(\mathbf{H}_{n,i}^{b}\right)^{H}\cdot\mathbf{H}_{n,i}^{b}\right|$, while the BF scheme performs the classical zero-force beamforming algorithm with equal power allocation. The FLOPs of this baseline algorithm can be given by $FLOPs_{\left(\textrm{GSA+ZFBF}\right)}\approx 6BNI\left[2M_{\textrm{t}}M_{\textrm{r}}+\left(2I+1\right)M_{\textrm{t}}M_{\textrm{r}}^{2}+\left(I^{2}+1\right)M_{\textrm{r}}^{3}\right]$.

Note that the RSA+ZFBF and GSA+ZFBF baselines are simple but commonly used in practical industrial scenarios. By comparing the proposed algorithms with these two baseline algorithms in terms of the sum rate and computational complexity under the 3GPP 38.901 UMa scenario, the practical application value of the proposed algorithms can be effectively demonstrated.

\subsection{Results and Discussions}

\begin{figure}[!t]
\centering
\includegraphics[width=0.95\columnwidth]{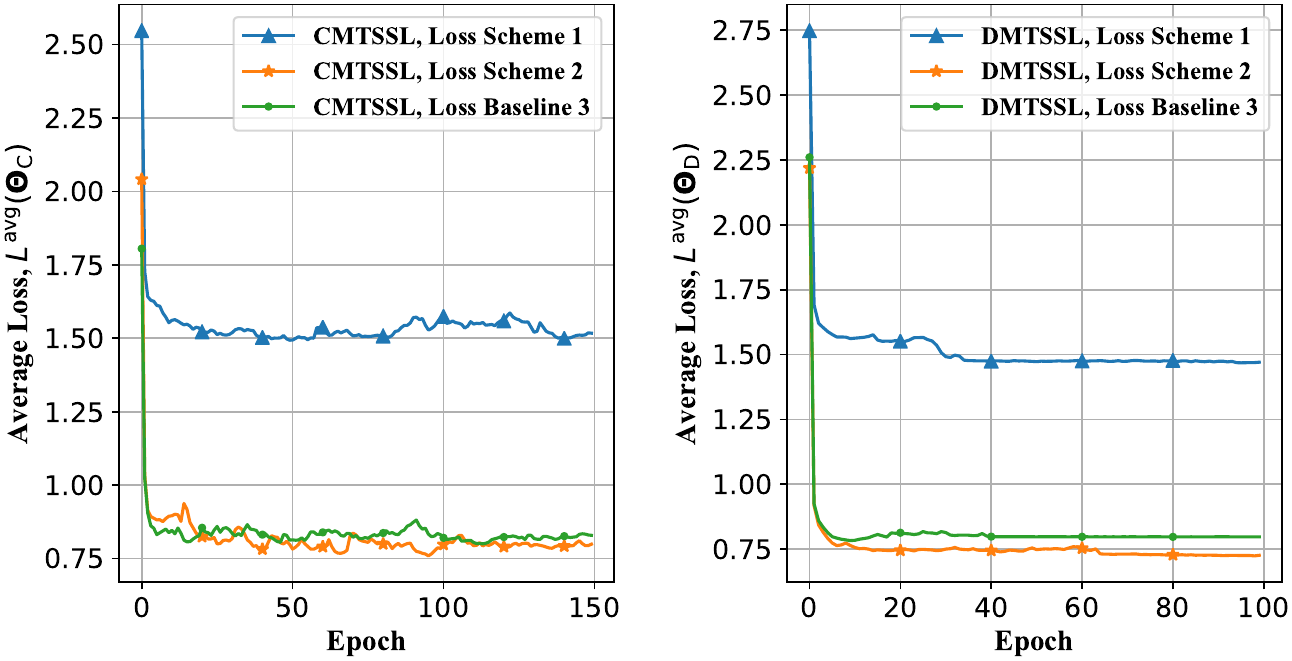}
\caption{Training losses of the proposed CMTSSL and DMTSSL algorithms on different loss schemes. ($B=3,I=10,N=4$)}
\label{Conver}
\end{figure}

Firstly, we use the proposed CMTSSL and DMTSSL algorithms to verify the effectiveness and superiority of the proposed loss functions. Specifically, by respectively applying the different loss schemes to the CMTSSL and DMTSSL algorithms, the performance comparisons between the proposed loss schemes and all the loss baselines are presented in Table \ref{Loss_eva}. It can be seen from Table \ref{Loss_eva} that the proposed loss schemes, i.e., the loss scheme 1 based on the proposed NFL loss and the loss scheme 2 based on the proposed EL loss, both significantly outperforms all the loss baselines, where the loss baseline 1 and 2 are even not convergent. This result shows the superiority and stability of the proposed NFL and EL losses in solving the joint optimization problem. Moreover, we can observe from Table \ref{Loss_eva} that, for the CMTSSL algorithm, the proposed loss scheme 1 is superior to the proposed loss scheme 2, while it is the opposite for the DMTSSL algorithm, which indicates that the proposed NFL loss is more suitable for the centralized self-supervised training while the proposed EL loss is more suitable for the distributed self-supervised training. In addition, the convergence behaviors of the proposed CMTSSL and DMTSSL algorithms in different loss schemes are illustrated in Fig. \ref{Conver}, where the loss curves of the proposed algorithms on loss baseline 1 and 2 are not shown in Fig. \ref{Conver} because they are not convergent. We can observe from Fig. \ref{Conver} that the proposed DMTSSL algorithm converges faster and stabler than the proposed CMTSSL algorithm. The reason is that the proposed DMTSSL algorithm consists of some distributed small NN models with low input/output dimensions while the proposed CMTSSL algorithm consists of a centralized large NN model with high input/output dimensions. For a similar task, the small NN model is commonly easier to train.

\begin{figure*}[!t]
\centering
\subfigure[]{\includegraphics[width=0.66\columnwidth]{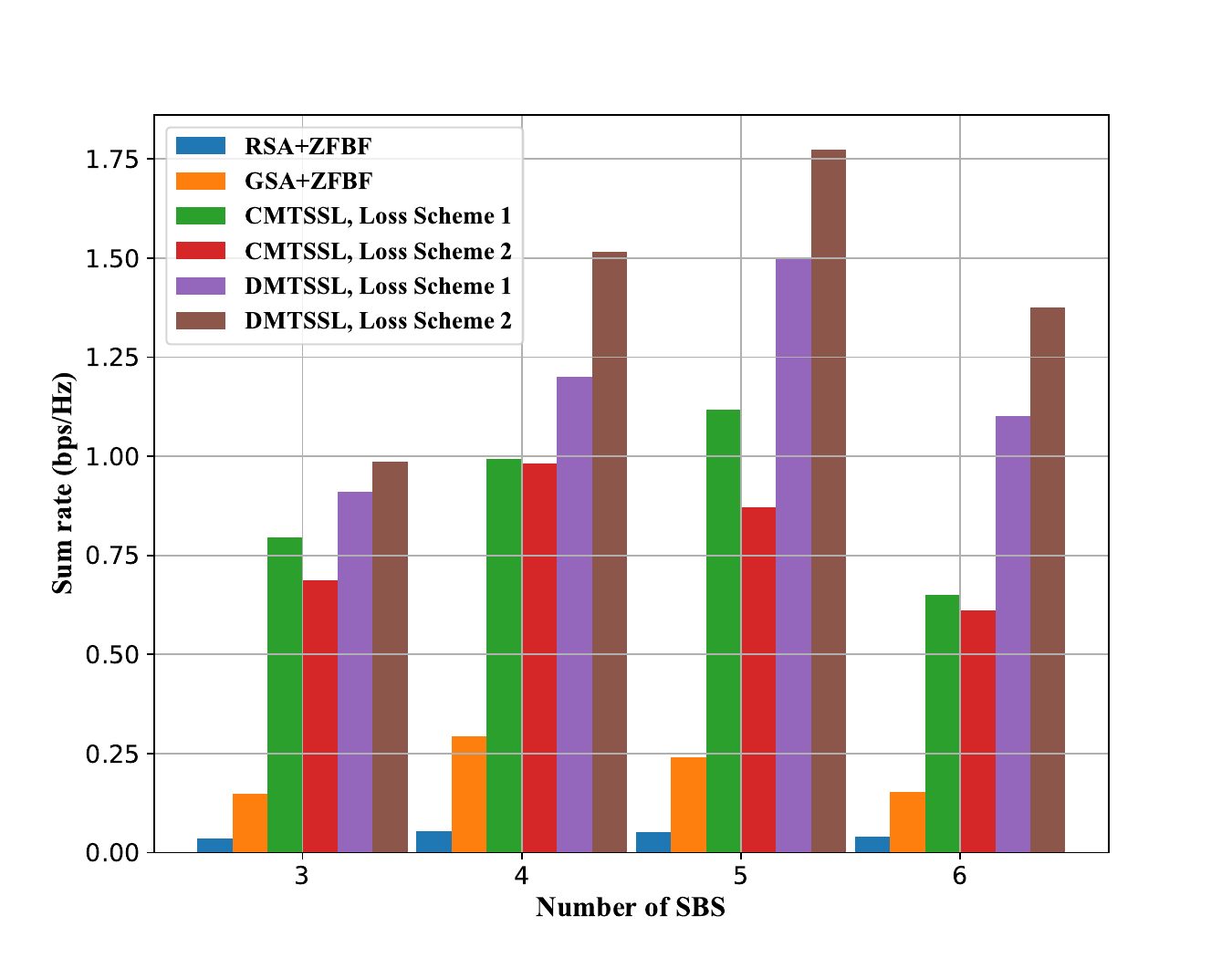}}
\subfigure[]{\includegraphics[width=0.66\columnwidth]{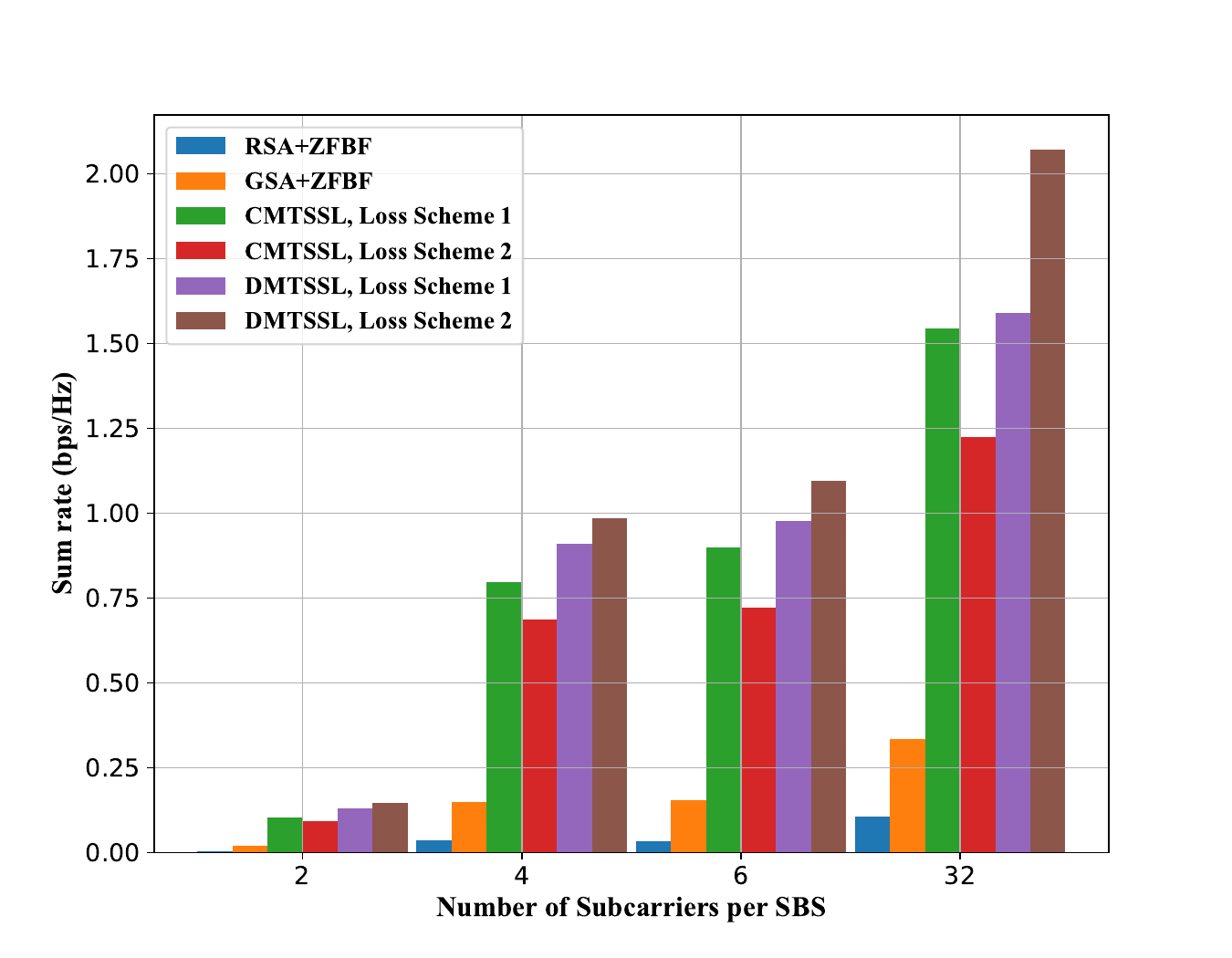}}
\subfigure[]{\includegraphics[width=0.66\columnwidth]{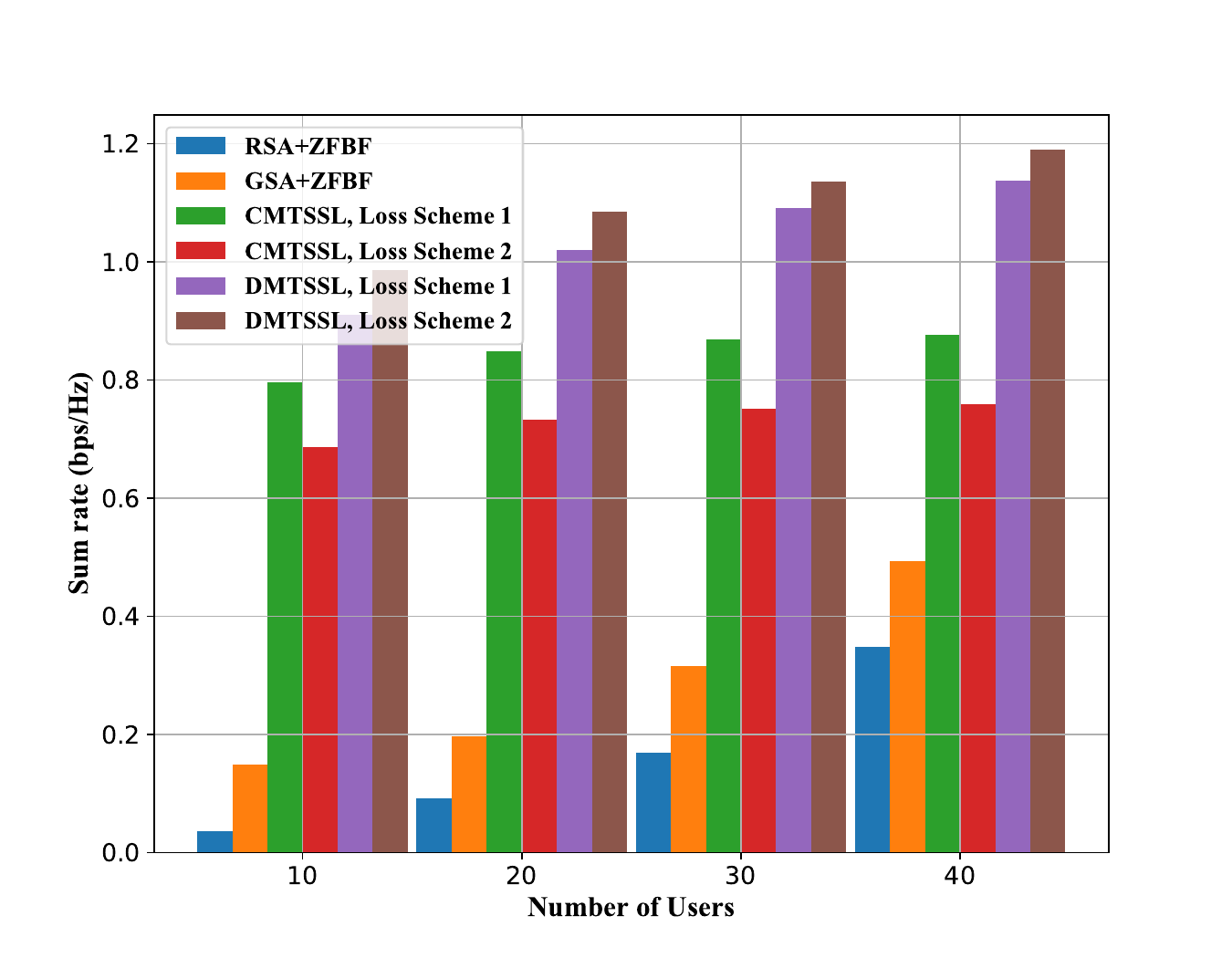}}
\caption{Performance comparisons between the proposed algorithms and the baseline algorithms. (a) Performance evaluations of the algorithms on different $B$. ($I=10,N=4$) (b) Performance evaluations of the algorithms on different $N$. ($B=3,I=10$) (c) Performance evaluations of the algorithms on different $I$. ($B=3,N=4$)}
\label{percompar}
\vspace{-1.5em}
\end{figure*}
\begin{figure*}[!t]
\centering
\subfigure[]{\includegraphics[width=0.66\columnwidth]{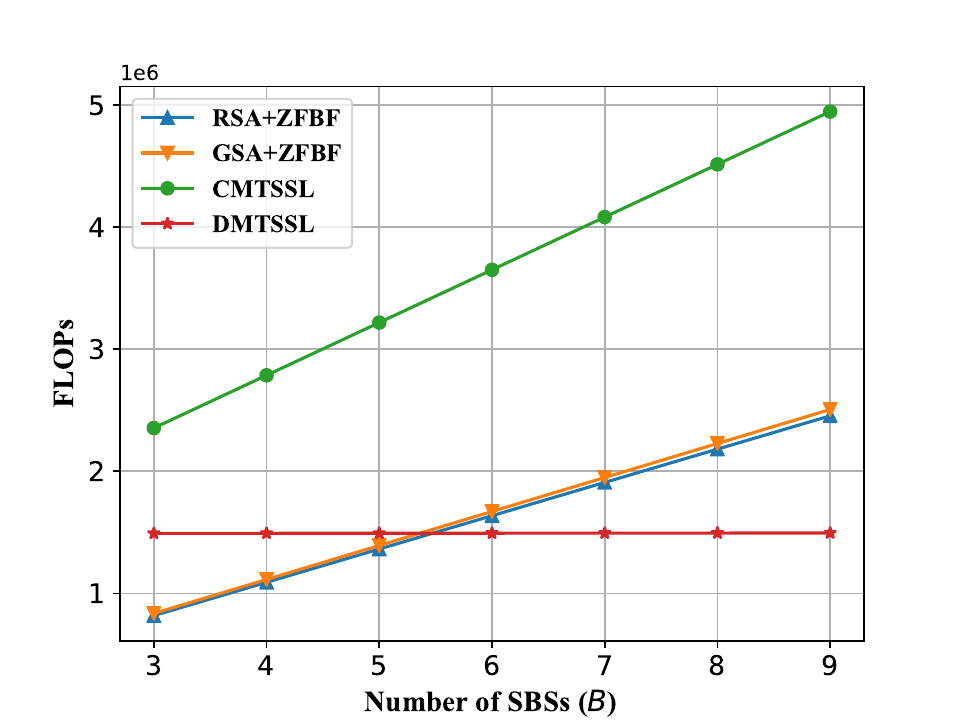}}
\subfigure[]{\includegraphics[width=0.66\columnwidth]{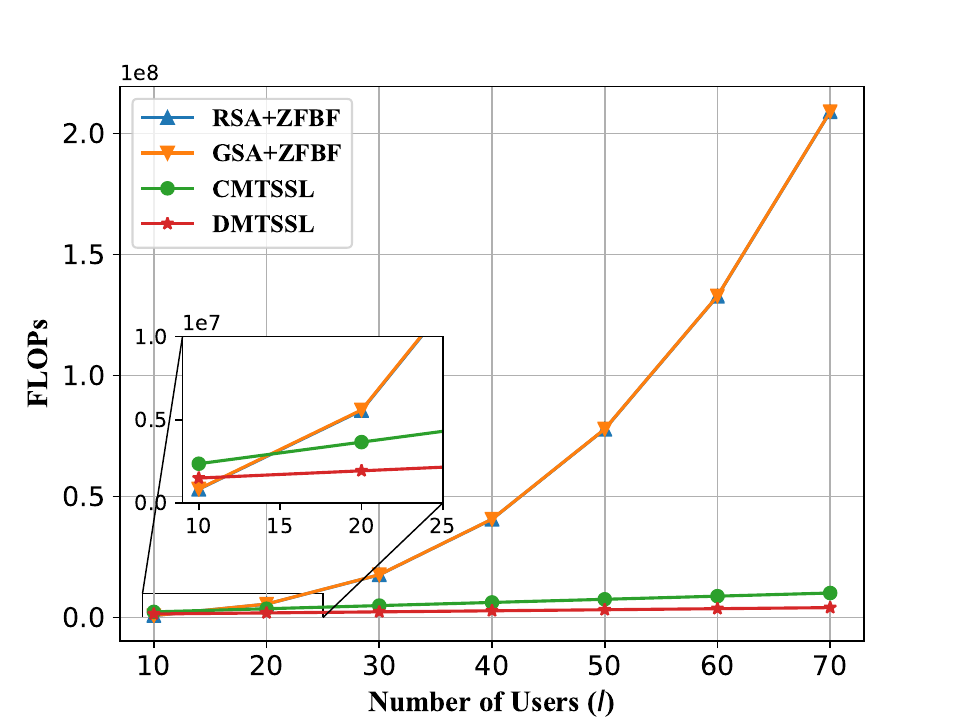}}
\subfigure[]{\includegraphics[width=0.66\columnwidth]{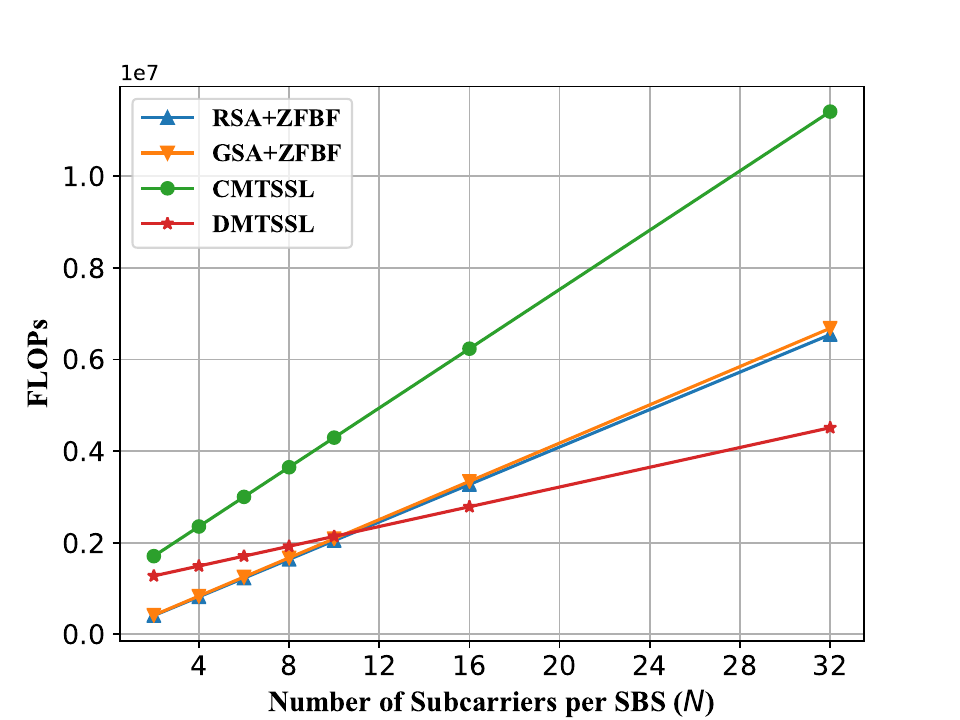}}
\caption{Complexity comparisons between the proposed algorithms and the baseline algorithms in terms of FLOPs. (a) Complexity of the algorithms on different $B$. ($I=10,N=4$) (b) Complexity of the algorithms on different $I$. ($B=3,N=4$) (c) Complexity of the algorithms on different $N$. ($B=3,I=10$)}
\label{complexity}
\vspace{-1.5em}
\end{figure*}

The performance comparisons among the proposed algorithms and all the baseline algorithms versus the number SBS $B$ are presented in Fig. \ref{percompar}(a). It can be seen from Fig. \ref{percompar}(a) that both the CMTSSL and DMTSSL algorithms significantly outperform all the baseline algorithms in terms of sum rate. Furthermore, Fig. \ref{percompar}(a) also shows that the proposed DMTSSL algorithm is superior to the proposed CMTSSL algorithm on the same loss scheme in terms of the sum rate regardless of the value of $B$. In addition, we can observe from Fig. \ref{percompar}(a) that the system sum rate tends to decrease as the number of SBSs increases to a certain extent. We believe that this is because users and SBSs are distributed in a small space with a fixed range in our simulation environment, while excessively high SBS density will cause intense interstation interference in the CF-MIMO communication scenario and then result in the decrease of the system sum rate.

In Fig. \ref{percompar}(b), we provide the sum-rate performance comparisons with respect to the number of subcarriers $N$. Fig. \ref{percompar}(b) indicates that both the proposed CMTSSL algorithm and the proposed DMTSSL algorithm significantly outperform all the baseline algorithms, while their individual sum rates increase with an increasing $N$. Likewise, the performance comparisons between the proposed algorithms and all the baseline algorithms versus the number of users $I$ are presented in Fig. \ref{percompar}(c). It can be seen from Fig. \ref{percompar}(c) that the proposed algorithms still outperform all the baselines regardless of the value of $I$.

We use the common metric, i.e., FLOPs, to measure the computational complexity of all the algorithms. The complexity comparisons between the proposed algorithms and the baseline algorithms are illustrated in Fig. \ref{complexity}, where the FLOPs of the baseline algorithms are calculated by the theoretical analysis while the FLOPs of the proposed algorithms are given by the actual statistics of the {\em Python} library. In addition, we count the FLOPs of one local NN model in the DMTSSL algorithm as the FLOPs of the DMTSSL algorithm due to the fact that all the local NN models in the proposed DMTSSL algorithm have the same structure and are independently distributed on different MEC servers. We can observe from Fig. \ref{complexity} that the complexities of the proposed CMTSSL algorithm and DMTSSL algorithm are linear functions with respect to the $B$, $I$ and $N$, while the complexities of the RSA+ZFBF and GSA+ZFBF baselines are power functions of the $I$. Particularly in Fig. \ref{complexity}(b), we can find that, with increasing $I$, the FLOPs of the proposed CMTSSL and DMTSSL algorithms are both significantly lower than that of the baseline algorithms. Fig. \ref{complexity}(a) and (c) indicate that the FLOPs of the proposed DMTSSL algorithm are still lower than that of the baseline algorithms at large $B$ or large $N$, while the FLOPs of the proposed CMTSSL algorithm become uncompetitive. It is worth to note that the proposed two algorithms are consistently superior to the baseline algorithms in terms of sum rate performance even if their complexities are higher than that of the baseline algorithms in some cases. Moreover, we can also observe from Fig. \ref{percompar} and Fig. \ref{complexity} that the proposed DMTSSL algorithm consistently outperforms the proposed CMTSSL algorithm in terms of both complexity and sum-rate performance regardless of the value variation of $B$, $I$ and $N$, which confirms the effectiveness and superiority of the proposed DMTSSL algorithm and is also consistent with our original intention of developing the DMTSSL algorithm \footnote{Note that many works \cite{ChenM20,LoweR17,Iqbal19} have also demonstrated that distributed learning algorithms, i.e., multi-agent reinforcement learning and federated learning, can outperform the centralized learning algorithms in some cases.}.

\begin{figure*}[!t]
\centering
\subfigure[]{\includegraphics[width=0.66\columnwidth]{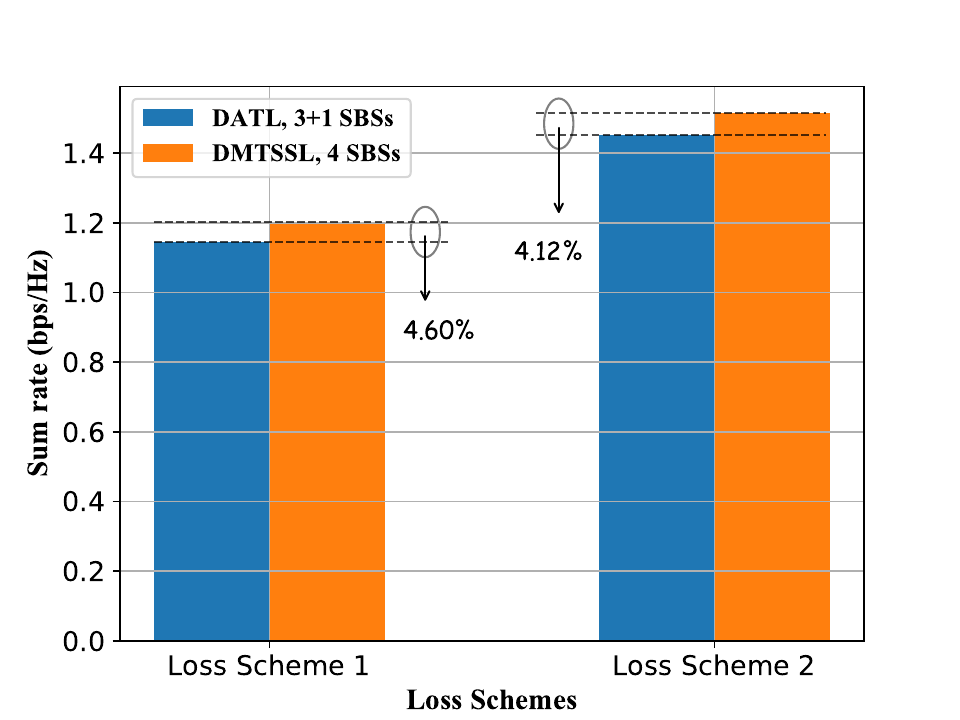}}
\subfigure[]{\includegraphics[width=0.66\columnwidth]{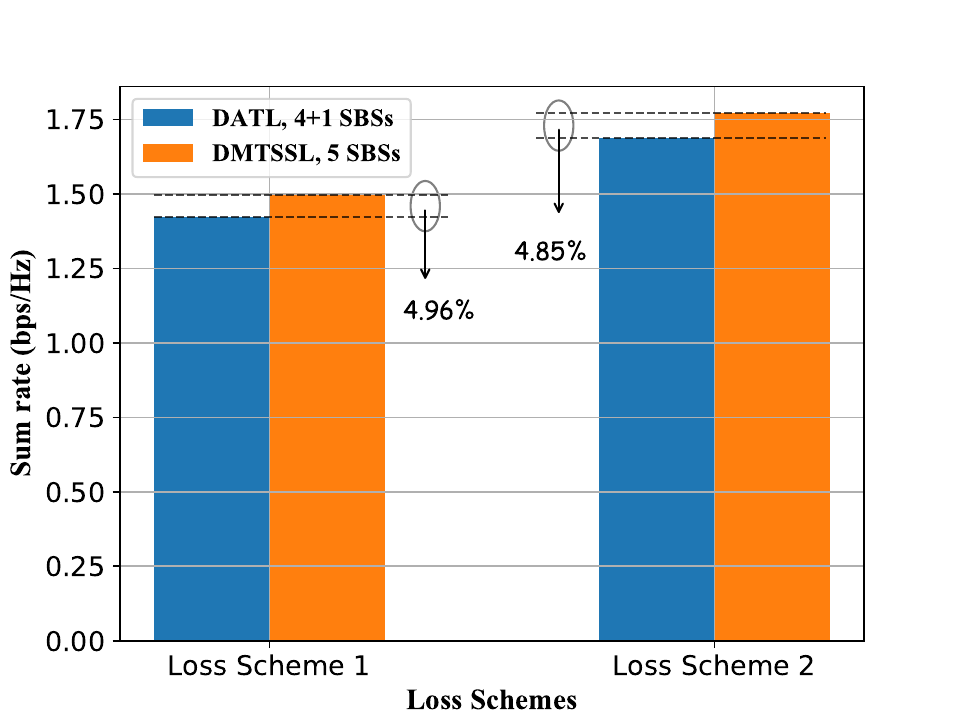}}
\subfigure[]{\includegraphics[width=0.66\columnwidth]{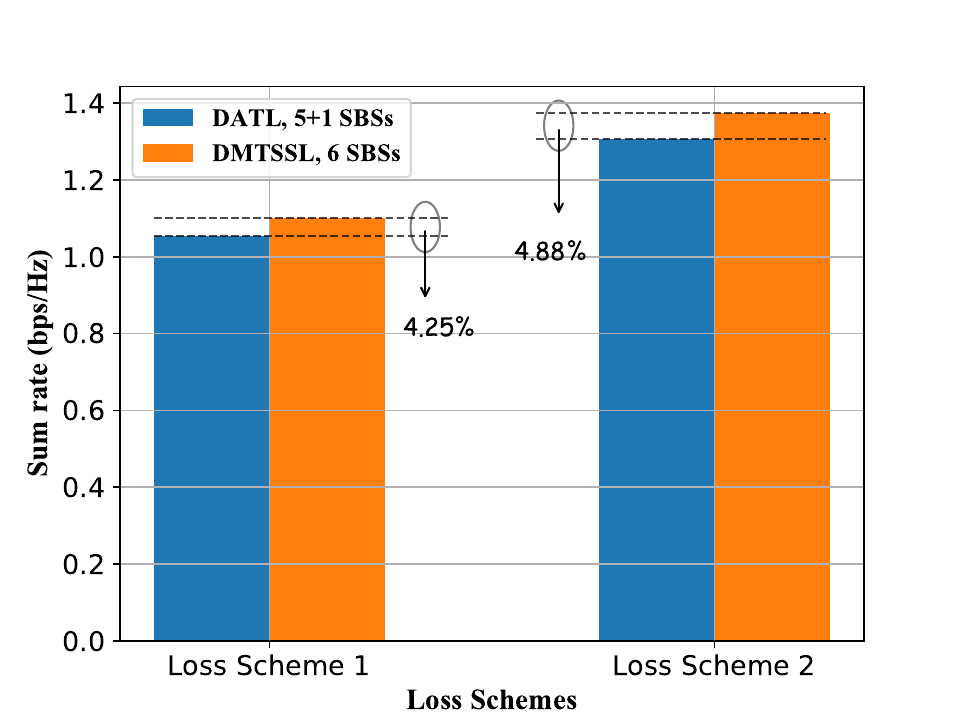}}
\caption{Performance evaluations of the proposed DATL algorithm in the dynamic scenarios with different number of original SBSs. ($I=10,N=4$) (a) Performance evaluation on 3 original SBSs and 1 new SBS. (b) Performance evaluation on 4 original SBSs and 1 new SBS. (c) Performance evaluation on 5 original SBSs and 1 new SBS.}
\label{DATL_diffB}
\vspace{-1.5em}
\end{figure*}

We take a step forward and study the sum-rate performance of the proposed DATL algorithm in the dynamic scenario where a new SBS is suddenly added to the original communication system. Concretely, the performance evaluations of the proposed DATL algorithm in the dynamic scenarios with different numbers of original SBSs are illustrated in Fig. \ref{DATL_diffB}. Due to the fact that the proposed DATL algorithm is developed based on the proposed DMTSSL algorithm, we thus set the DMTSSL algorithm as the comparison, i.e., we retrain the DMTSSL algorithm when there is a new SBS added and then obtain the sum-rate performance of the retrained DMTSSL algorithm in the new communication system. Finally, Fig. \ref{DATL_diffB} shows that the proposed DATL algorithm can avoid extra costs arising from the NN retraining or any other complex calculations, while its performance loss in terms of the sum rate is less than 5\% of the retrained DMTSSL algorithm.

\section{Conclusion}\label{sec6}

In this paper, the joint SA and BF cross-layer resource allocation problem in the MEC-aided OFDMA CF-MIMO communication network is formulated to maximize the weighted sum rate while subjecting to the requirements of pre-user minimum tolerant data rate and pre-SBS maximum allowable transmitting power. We designed two novel and general self-supervised loss functions, i.e., NFL loss and EL loss, and proposed three learning algorithms, namely CMTSSL, DMTSSL, and DATL, to effectively solve the cross-layer resource allocation problem while avoiding the technical bottlenecks of traditional methods, e.g., the non-convexity, the complexity disaster and the NP-hard issue, as well as addressing some open challenges faced by the learning-based methods, e.g., the costly manual labeling, the dimension disaster for massive output, and the scalability limitations. The superior performance of the proposed algorithms compared to the baseline algorithms is confirmed by numerical simulations.

\appendices
\section{ANALYSIS OF THE NFL AND EL LOSSES}\label{apend_A}

\subsection{Robustness Analysis}

We can analyze the robustness of the designed loss functions, i.e., NFL loss and EL loss, by applying these losses to a general multi-class classification problem. Note that, a loss function $L$ is robust to label noise if $L$ is noice-tolerant to noisy labels \cite{Ghosh17}. For a general $Z$-class classification problem, we assume standard neural network architecture with softmax output layer and mean-absolute error residual calculation. If the network input is $\mathbf{x}$, then the input of the softmax layer can be denoted as $G\left(\mathbf{x}\right)$. The softmax layer computes
\begin{sequation}\label{eq_29}
u_{z}=\frac{{\rm exp}\left(G\left(\mathbf{x}\right)_{z}\right)}{\sum_{z=1}^{Z}{\rm exp}\left(G\left(\mathbf{x}\right)_{z}\right)},z\in\left\{ 1,2,\cdots,Z\right\}
\end{sequation}
\hspace{-4pt}where $G\left(\mathbf{x}\right)_{z}$ represents the $z^{\rm th}$ componet of $G\left(\mathbf{x}\right)$. We define the output of the softmax layer as $\mathbf{u}=F\left(\mathbf{x}\right)=\left[u_{1},\cdots,u_{Z}\right]$. $F$ is the classifier. From \eqref{eq_29}, we can easily have $\sum_{z=1}^{Z}u_{z}=1$. Moreover, if the class of $\mathbf{x}$ is $c\in\left\{ 1,\cdots,Z\right\} $ then the label $y_{\mathbf{x}}$, which is in `one-of-$Z$' representation, is given as $\mathbf{q}_{c}=\left[q_{1c},\cdots,q_{Zc}\right]$ where $q_{zc}=1$ when $z=c$, otherwise 0. Assume that the label is polluted to $\hat{y}_{\mathbf{x}}$ by symmetric or uniform noise with a noise rate of $\eta$. We have
\begin{sequation}
\hat{y}_{\mathbf{x}}=\begin{cases}
y_{\mathbf{x}} & \textrm{with\;prbobality\;}\left(1-\eta\right)\\
\mathbf{q}_{j},j\in\left[z\right],j\neq c & \textrm{with\;prbobality\;}\eta_{j}
\end{cases}
\end{sequation}
\hspace{-4pt}where $\sum_{j\neq c}\eta_{j}=\eta$. Note that $\eta_{i}=\eta_{j},\forall i,j\in\left[z\right],i\neq c,j\neq c$ is satisfied when the noise is symmetric or uniform.

A loss function $L$ is noise-tolerant if the minimizer of risk under $L$ with noisy labels would be the same as that with noise-free labels \cite{Ghosh17}. Therein, the risk under $L$ can be defined by 
\begin{sequation}
R_{L}\left(F\right)=\mathbb{E}_{\mathbf{x},y_{\mathbf{x}}}\left[L\left(F\left(\mathbf{x}\right),y_{\mathbf{x}}\right)\right].
\end{sequation}
\hspace{-4pt}$\mathbb{E}$ denotes expectation with respect to the variables or distributions indicated by the subscript.

After that, we give the following Theorem \ref{theorem1} to prove the noice-tolerant property of the designed NFL and EL losses.
\begin{theorem}\label{theorem1}
In a general $Z$-class classification problem, the NFL loss and EL loss are both noice-tolerant to symmetric or uniform label noise when the noise rate $\eta<\frac{Z-1}{Z}$ as well as the loss input $x>x_1$ for the NFL loss or $x>x_2$ for the EL loss.
\end{theorem}
\begin{IEEEproof}
For symmetric or uniform noise, the risk under $L$ with noisy labels, denoted by $R_{L}^{\eta}\left(F\right)$, can be derived as \eqref{R_n_L}.

\begin{figure*}[!htbp]
\newcounter{MYtempeqncnt}
\setcounter{MYtempeqncnt}{\value{equation}}
\setcounter{equation}{28}
{\small \begin{align}\label{R_n_L}
R_{L}^{\eta}\left(F\right)&=\mathbb{E}_{\mathbf{x},\hat{y}_{\mathbf{x}}}\left[L\left(F\left(\mathbf{x}\right),\hat{y}_{\mathbf{x}}\right)\right]=\mathbb{E}_{\mathbf{x}}\mathbb{E}_{\left.y_{\mathbf{x}}\right|\mathbf{x}}\mathbb{E}_{\left.\hat{y}_{\mathbf{x}}\right|\mathbf{x},y_{\mathbf{x}}}\left[L\left(F\left(\mathbf{x}\right),\hat{y}_{\mathbf{x}}\right)\right] \nonumber \\
&=\mathbb{E}_{\mathbf{x}}\mathbb{E}_{\left.y_{\mathbf{x}}\right|\mathbf{x}}\left[\left(1-\eta\right)L\left(F\left(\mathbf{x}\right),y_{\mathbf{x}}\right)+\frac{\eta}{Z-1}\sum_{j\neq c}L\left(F\left(\mathbf{x}\right),\mathbf{q}_{j}\right)\right] \nonumber \\
&=\left(1-\eta\right)R_{L}\left(F\right)+\frac{\eta}{Z-1}\mathbb{E}_{\mathbf{x},y_{\mathbf{x}}}\left[\sum_{j=1}^{Z}L\left(F\left(\mathbf{x}\right),\mathbf{q}_{j}\right)-L\left(F\left(\mathbf{x}\right),y_{\mathbf{x}}\right)\right] \nonumber \\
&=\left(1-\frac{\eta\cdot Z}{Z-1}\right)R_{L}\left(F\right)+\frac{\eta}{Z-1}\mathbb{E}_{\mathbf{x},y_{\mathbf{x}}}\left[\sum_{j=1}^{Z}L\left(F\left(\mathbf{x}\right),\mathbf{q}_{j}\right)\right].
\end{align}}
\begin{sequation}\label{loss_zclass}
L\left(F\left(\mathbf{x}\right),\mathbf{q}_{j}\right)=\begin{cases}
\begin{cases}
\frac{1}{x_{1}^{2}}\left(\left\Vert \mathbf{q}_{j}-\mathbf{u}\right\Vert _{1}-x_{1}\right)=\frac{1}{x_{1}^{2}}\left(2-x_{1}-2u_{j}\right), & \left\Vert \mathbf{q}_{j}-\mathbf{u}\right\Vert _{1}\geq x_{1}\\
-\frac{1}{\left\Vert \mathbf{q}_{j}-\mathbf{u}\right\Vert _{1}}=-\frac{1}{2-2u_{j}}, & \left\Vert \mathbf{q}_{j}-\mathbf{u}\right\Vert _{1}<x_{1}<0
\end{cases} & \textrm{NFL}\\
\begin{cases}
e^{x_{2}}\left(\left\Vert \mathbf{q}_{j}-\mathbf{u}\right\Vert _{1}+1-x_{2}\right)=e^{x_{2}}\left(3-x_{2}-2u_{j}\right), & \left\Vert \mathbf{q}_{j}-\mathbf{u}\right\Vert _{1}\geq x_{2}\\
e^{\left\Vert \mathbf{q}_{j}-\mathbf{u}\right\Vert _{1}}=e^{2-2u_{j}}, & \left\Vert \mathbf{q}_{j}-\mathbf{u}\right\Vert _{1}<x_{2}
\end{cases} & \textrm{EL}
\end{cases}.
\end{sequation}
{\small \begin{align}\label{R_n_L_final}
R_{L}^{\eta}\left(F\right)&=\left(1-\frac{\eta\cdot Z}{Z-1}\right)R_{L}\left(F\right)+\begin{cases}
\frac{\eta}{Z-1}\mathbb{E}_{\mathbf{x},y_{\mathbf{x}}}\left[\sum_{j=1}^{Z}\frac{1}{x_{1}^{2}}\left(2-x_{1}-2u_{j}\right)\right], & \textrm{NFL}\\
\frac{\eta}{Z-1}\mathbb{E}_{\mathbf{x},y_{\mathbf{x}}}\left[\sum_{j=1}^{Z}e^{x_{2}}\left(3-x_{2}-2u_{j}\right)\right], & \textrm{EL}
\end{cases} \nonumber \\
&=\left(1-\frac{\eta\cdot Z}{Z-1}\right)R_{L}\left(F\right)+\begin{cases}
\frac{\eta}{Z-1}\cdot\frac{2\left(Z-1\right)-Zx_{1}}{x_{1}^{2}}, & \textrm{NFL}\\
\frac{\eta}{Z-1}\cdot\left(3Z-2-Zx_{2}\right)e^{x_{2}}, & \textrm{EL}
\end{cases} \nonumber \\
&=\left(1-\frac{\eta\cdot Z}{Z-1}\right)R_{L}\left(F\right)+\begin{cases}
C_{\textrm{NFL}}\cdot\frac{\eta}{Z-1}, & \textrm{NFL}\\
C_{\textrm{EL}}\cdot\frac{\eta}{Z-1}, & \textrm{EL}
\end{cases},
\end{align}}
\setcounter{equation}{\value{MYtempeqncnt}}
\hrulefill
\end{figure*}

In this general $Z$-class classification problem, $L\left(F\left(\mathbf{x}\right),\mathbf{q}_{j}\right)$ under the NFL loss and EL loss can be written as \eqref{loss_zclass}.

When the loss input $x=\left\Vert \mathbf{q}_{j}-\mathbf{u}\right\Vert _{1}\geq x_{1}$ for the NFL loss or $x=\left\Vert \mathbf{q}_{j}-\mathbf{u}\right\Vert _{1}\geq x_{2}$ for the EL loss, the $R_{L}^{\eta}\left(F\right)$ can be further derived as \eqref{R_n_L_final}, 
where both $C_{\textrm{NFL}}$ and $C_{\textrm{NFL}}$ are constants. Then, for any $F$, let $F^{*}$ be the minimizer of $R_{L}\left(F\right)$. We have
\setcounter{equation}{31}
\begin{sequation}
R_{L}^{\eta}\left(F^{*}\right)-R_{L}^{\eta}\left(F\right)=\left(1-\frac{\eta\cdot Z}{Z-1}\right)\left(R_{L}\left(F^{*}\right)-R_{L}\left(F\right)\right)\leq0,
\end{sequation}
\hspace{-3pt}when $\eta<\frac{Z-1}{Z}$ is satisfied. This prove $F^{*}$ is also the minimizer of $R_{L}^{\eta}\left(F\right)$. Therefore, the NFL loss is robust to symmetric or uniform label noise when the noise rate $\eta<\frac{Z-1}{Z}$ and the loss input $x>x_1$. Similarly, the EL loss is also robust to symmetric or uniform label noise when the noise rate $\eta<\frac{Z-1}{Z}$ and the loss input $x>x_2$.

\end{IEEEproof}

\vspace{-4pt}
\subsection{Target Domain Discussions}

\begin{figure}[!b]
\centering
\subfigure[]{\includegraphics[width=0.48\columnwidth]{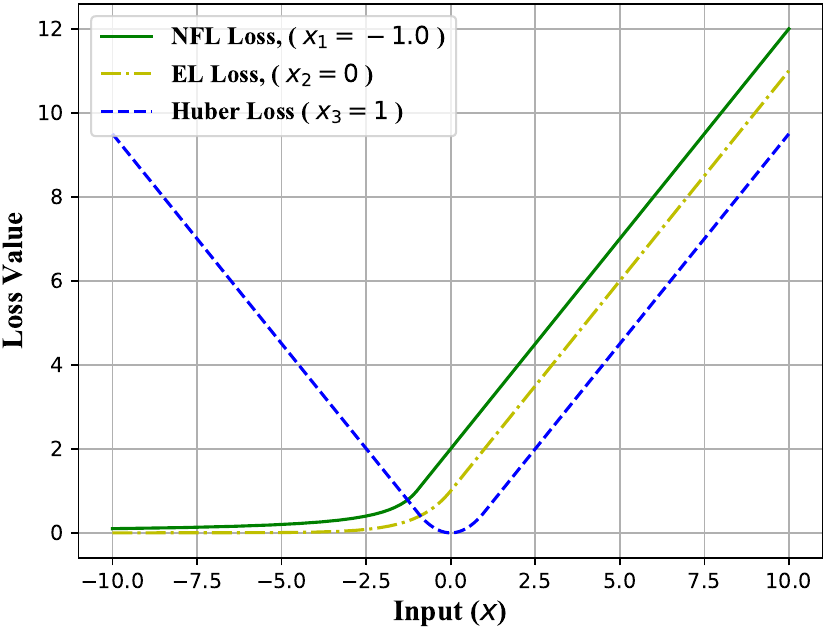}}
\subfigure[]{\includegraphics[width=0.48\columnwidth]{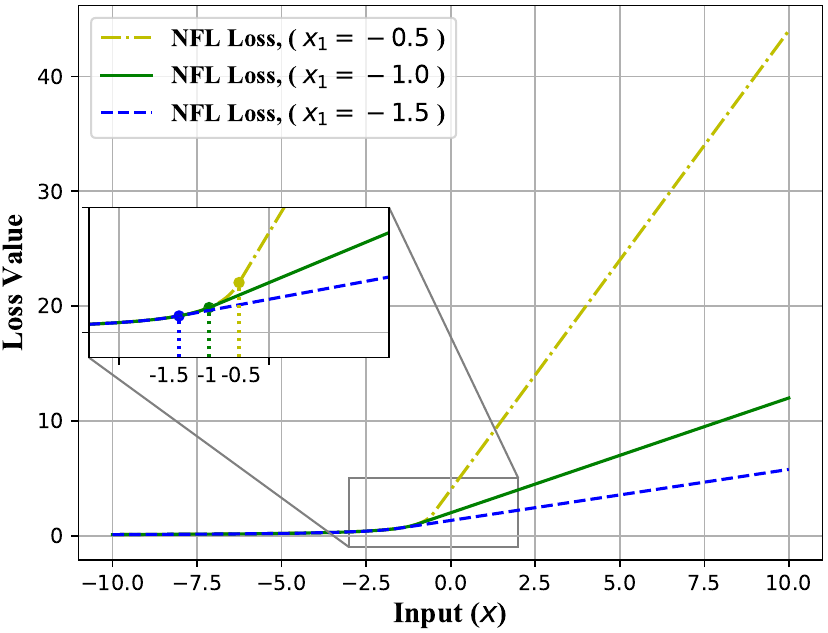}}
\subfigure[]{\includegraphics[width=0.48\columnwidth]{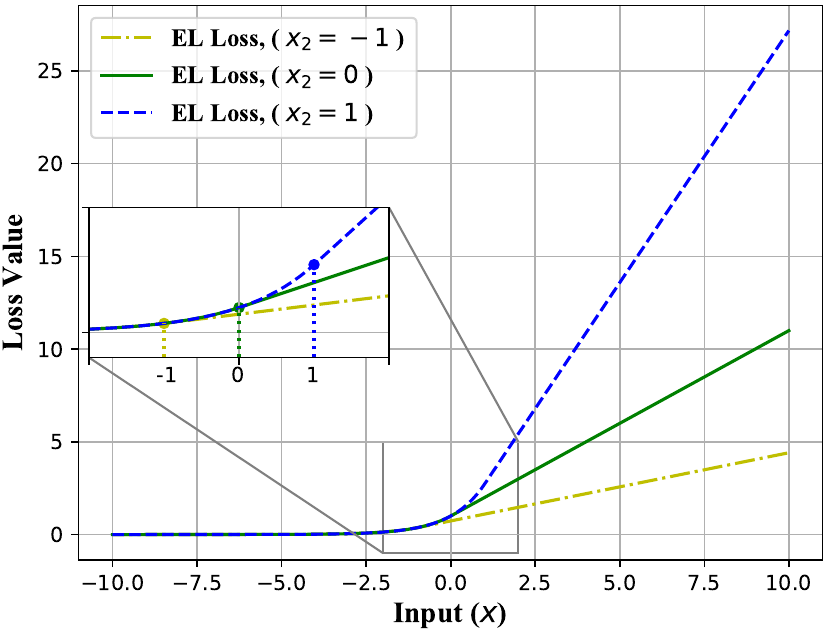}}
\subfigure[]{\includegraphics[width=0.48\columnwidth]{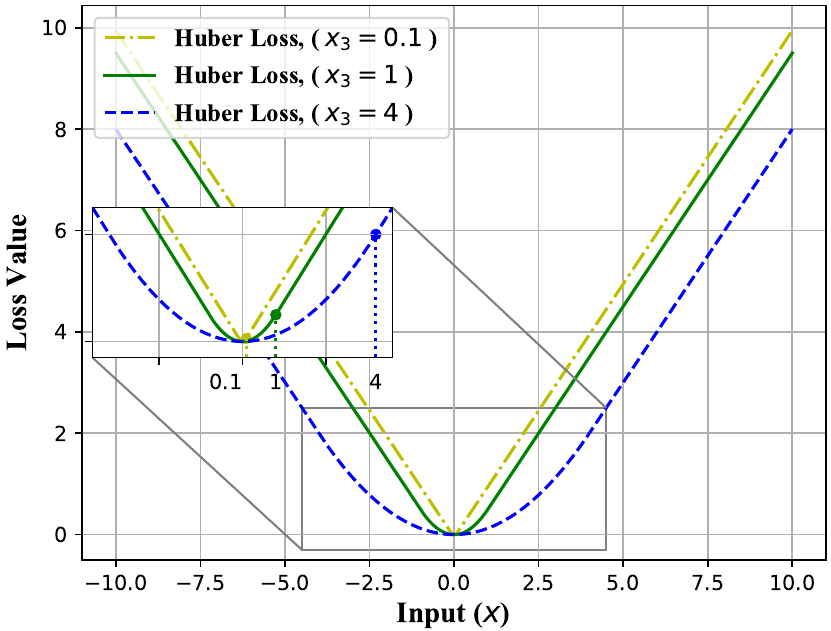}}
\caption{Loss functions under different hyper-parameters. (a) Three loss functions at $x_1=-1$, $x_2=0$, and $x_3=1$, respectively. (b) NFL loss at different $x_1$. (c) EL loss at different $x_2$. (b) Huber loss at different $x_3$.}
\label{designLoss}
\vspace{-1.5em}
\end{figure}
The target domain of a self-supervised learning task should be the same as the convergence domain of the loss function input when the loss function output approaches zero. These three loss functions under different hyper-parameters are illustrated in Fig. \ref{designLoss} for clearer observation. From Fig. \ref{designLoss}, we can find a common characteristic between the NFL loss and EL loss that the loss input $x$ gradually approaches minus infinity with a decreasing loss value. Therefore, the NFL and EL losses are suitable for tasks whose target domain is the negative infinite domain. Likewise, we can also observe from Fig. \ref{designLoss} that the Huber loss is suitable for those tasks whose target value is expected to converge to zero.

\section{DERIVATION OF THE JOINT LOSS}\label{apend_B}
Let $F_{\mathbf{\Theta}}\left(\mathbf{x}\right)$ denote the output of a genneral NN with weights $\mathbf{\Theta}$ on input $\mathbf{x}$. The NN can be viewed as a probabilistic model and the likelihood can be defined as a Gaussian with mean given by the model output $\mathbf{y}$ and with an observation noise scalar denoted as $\beta$. Specifically,
{\small \begin{align}\label{proba_model}
p\left(\left.\mathbf{y}\right|F_{\mathbf{\Theta}}\left(\mathbf{x}\right)\right)=\mathcal{N}\left(F_{\mathbf{\Theta}}\left(\mathbf{x}\right),\beta^{2}\right).
\end{align}}
\hspace{-5pt}Note that $\beta$ captures the intensity of noise in the model output. For the $K$-task learning with multiple model outputs $\mathbf{y}_{1},\cdots,\mathbf{y}_{K}$, given sufficient statistics, we can define the likelihood to factorize over the outputs. By define $F_{\mathbf{\Theta}}\left(\mathbf{x}\right)$ as the sufficient statistics, we obtain the multi-task likelihood
{\small \begin{align}
p\left(\left.\mathbf{y}_{1},\cdots,\mathbf{y}_{K}\right|F_{\mathbf{\Theta}}\left(\mathbf{x}\right)\right)=p\left(\left.\mathbf{y}_{1}\right|F_{\mathbf{\Theta}}\left(\mathbf{x}\right)\right)\cdots p\left(\left.\mathbf{y}_{K}\right|F_{\mathbf{\Theta}}\left(\mathbf{x}\right)\right),
\end{align}}
\hspace{-5pt}where $K \geq 2$. Then, we define the joint loss $L_{\mathbf{\Theta}}\left(\mathbf{y}_{1},\cdots,\mathbf{y}_{K},\beta_{1},\cdots,\beta_{K}\right)$ for the multi-output model by minimizing the negative log likelihood:
{\small \begin{align}\label{org_joinloss}
&=-\textrm{log}\left(p\left(\left.\mathbf{y}_{1},\cdots,\mathbf{y}_{K}\right|F_{\mathbf{\Theta}}\left(\mathbf{x}\right)\right)\right) \nonumber \\
&=-\textrm{log}\left(p\left(\left.\mathbf{y}_{1}\right|F_{\mathbf{\Theta}}\left(\mathbf{x}\right)\right)\cdots p\left(\left.\mathbf{y}_{K}\right|F_{\mathbf{\Theta}}\left(\mathbf{x}\right)\right)\right) \nonumber \\
&=-\textrm{log}\left(\mathcal{N}\left(\mathbf{y}_{1};F_{\mathbf{\Theta}}\left(\mathbf{x}\right),\beta_{1}^{2}\right)\cdots\mathcal{N}\left(\mathbf{y}_{K};F_{\mathbf{\Theta}}\left(\mathbf{x}\right),\beta_{K}^{2}\right)\right) \nonumber \\
&\propto\frac{1}{2\beta_{1}^{2}}\left\Vert \mathbf{y}_{1}-F_{\mathbf{\Theta}}\left(\mathbf{x}\right)\right\Vert ^{2}+\textrm{log}\beta_{1}+\cdots \nonumber\\
&+\frac{1}{2\beta_{K}^{2}}\left\Vert \mathbf{y}_{K}-F_{\mathbf{\Theta}}\left(\mathbf{x}\right)\right\Vert ^{2}+\textrm{log}\beta_{K} \nonumber \\
&=\sum_{k=1}^{K}\frac{1}{2\beta_{k}^{2}}\left\Vert \mathbf{y}_{k}-F_{\mathbf{\Theta}}\left(\mathbf{x}\right)\right\Vert ^{2}+\textrm{log}\left(\prod_{k=1}^{K}\beta_{k}\right),
\end{align}}
\hspace{-4pt}where $\beta_{k}$ is the noise parameter for the output $\mathbf{y}_{k}$. Let $L_{\mathbf{\Theta}}\left(\mathbf{y}_{k}\right)=\left\Vert \mathbf{y}_{k}-F_{\mathbf{\Theta}}\left(\mathbf{x}\right)\right\Vert ^{2}$ denote the loss of the $k$-th learning task. Moreover, to avoid convergence difficulties caused by excessive loss values in practical training, we multiply the sum of losses by a factor of $\frac{2}{K}$. Then, 
{\small \begin{align}\label{mid_joinloss}
L_{\mathbf{\Theta}}\left(\mathbf{y}_{1},\cdots,\mathbf{y}_{K},\beta_{1},\cdots,\beta_{K}\right)= \nonumber\\ 
\sum_{k=1}^{K}\frac{1}{K\beta_{k}^{2}}L_{\mathbf{\Theta}}\left(\mathbf{y}_{k}\right)+\textrm{log}\left(\prod_{k=1}^{K}\beta_{k}\right).
\end{align}}
\hspace{-4pt}$\beta_{k}$ can be further interpreted as the relative weight of the loss $L_{\mathbf{\Theta}}\left(\mathbf{y}_{k}\right)$, which can be adaptively learned based on the data. As $\beta_{k}$ increases, which means the noise for the output $\mathbf{y}_{k}$ increases and correspondingly the value of loss $L_{\mathbf{\Theta}}\left(\mathbf{y}_{k}\right)$ increase, we have that the weight of $L_{\mathbf{\Theta}}\left(\mathbf{y}_{k}\right)$ in the joint loss decreases. The last term in \eqref{mid_joinloss} can act as a regularizer for the noise terms to discourage the noise from increasing too high. Finally, we apply \eqref{mid_joinloss} into our multi-task learning problem and then obtain the final joint loss formulated in \eqref{Jointloss}.

\ifCLASSOPTIONcaptionsoff
  \newpage
\fi

\balance

\end{document}